%% file: main.tex
\documentclass[10pt,twocolumn,letterpaper]{article}

\usepackage{cvpr}              

\input{preamble}

\definecolor{cvprblue}{rgb}{0.21,0.49,0.74}
\usepackage[pagebackref,breaklinks,colorlinks,allcolors=cvprblue]{hyperref}

\def\paperID{43177} 
\def\confName{CVPR}
\def\confYear{2026}

\title{\logo Uni-DAD: Unified Distillation and Adaptation of Diffusion Models \\ for Few-step Few-shot Image Generation}

\author{
Yara~Bahram\thanks{Equal contribution},~
Mélodie~Desbos\footnotemark[1],~
Mohammadhadi~Shateri,~
Eric~Granger\\
LIVIA, ILLS, ETS Montreal, Canada\\
{\tt\small \{yara.mohammadi-bahram,melodie.desbos\}@livia.etsmtl.ca,}\\ 
{\tt\small \{mohammadhadi.shateri,eric.granger\}@etsmtl.ca}
}

\begin{document}
\maketitle

\input{sec/0_abstract}

\input{sec/1_intro}
\input{sec/2_related}

\input{sec/3_method}

\input{sec/4_results}
\input{sec/5_conclusion}

{
\small
\bibliographystyle{ieeenat_fullname}
\bibliography{main}
}
\input{sec/X_suppl} 


\end{document}

%% file: preamble.tex
\usepackage[table]{xcolor}

\newcommand{\tboxtext}[2]{
  {\setlength{\fboxsep}{1pt}\colorbox[HTML]{#1}{\strut #2}}%
}

\usepackage{capt-of} 
\usepackage{cuted}   

\usepackage{booktabs,multirow}
\usepackage{booktabs,tabularx}
\usepackage{tablefootnote}
\usepackage{amssymb}

\usepackage{pifont}
\newcommand{\cmark}{\ding{51}}

\newcommand{\fid}{FID$\downarrow$}
\newcommand{\na}{\textemdash}

\usepackage{graphicx}  

\usepackage{xcolor}
\usepackage[svgnames]{xcolor}
\definecolor{MyLime}{RGB}{128,192,0}

\definecolor{darkmaroon}{HTML}{3C0006}

\newlength{\colw}
\newcolumntype{C}[1]{>{\centering\arraybackslash}m{#1}}
\newcolumntype{L}[1]{>{\raggedright\arraybackslash}m{#1}}
\usepackage{tabularx,array,booktabs,multirow,xcolor}
\newcolumntype{Y}{>{\centering\arraybackslash}X}        
\newcolumntype{Z}{>{\centering\arraybackslash}X}

\DeclareUnicodeCharacter{202F}{ }

\usepackage{array}
\usepackage{booktabs}
\usepackage[table]{xcolor}
\usepackage{graphicx}
\usepackage{amssymb}

\usepackage{iftex}
\ifPDFTeX
\else
  \usepackage{fontspec}         
  \usepackage{newtxmath}
\fi

\usepackage{soul,xcolor} 

\newcommand{\outlinebox}[1]{
\begingroup 
\setlength{\fboxsep}{0.1ex}
\setlength{\fboxrule}{1.3pt}
\fcolorbox[HTML]{FF0000}{FFFFFF}{\strut #1}%
\endgroup 
}
\newcommand{\outlineboxx}[1]{
\begingroup 
\setlength{\fboxsep}{0.1ex}
\setlength{\fboxrule}{1.3pt}
\fcolorbox[HTML]{B3B3B3}{FFFFFF}{\strut #1}%
\endgroup 
}

\newcommand{\outlineboxxx}[1]{
\begingroup 
\setlength{\fboxsep}{0.1ex}
\setlength{\fboxrule}{1.3pt}
\fcolorbox[HTML]{D6B656}{FFFFFF}{\strut #1}%
\endgroup 
}

\definecolor{motiv_color1}{HTML}{FF0000}
\definecolor{motiv_color2}{HTML}{B3B3B3}
\definecolor{motiv_color3}{HTML}{D6B656}

\usepackage[ruled,vlined,linesnumbered]{algorithm2e} 

\usepackage{amsmath,amssymb,amsfonts,mathtools}

\usepackage{placeins}

\usepackage{graphicx}

\DeclareRobustCommand{\logo}{%
  \begingroup\normalfont
  \raisebox{-1.0ex}{\hspace{-0.35em}\includegraphics[height=5.5ex]{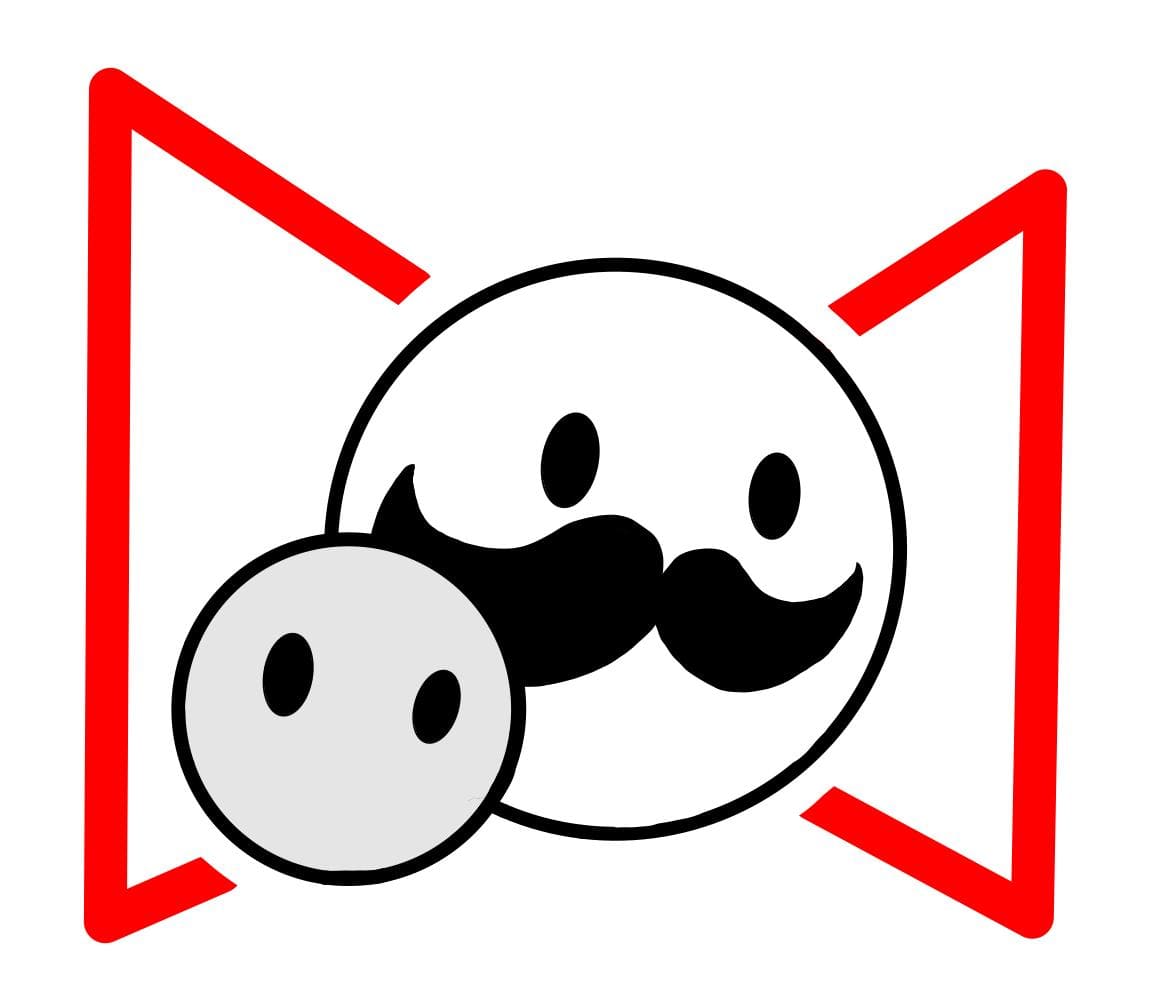}}%
  \kern0.3em
  \endgroup
}

\usepackage{amsmath,amssymb}
\usepackage[ruled,vlined,linesnumbered]{algorithm2e} 
\usepackage{xcolor} 
\newlength{\AssignLHS}   
\newlength{\AssignGap}   
\newlength{\ArrowW}      
\newlength{\EqTab}       

\SetKwComment{rtab}{}{} 
\newcommand{\eqmark}[1]{%
  \rtab*{%
    \textcolor{black}{//~Eq~}%
    \hyperref[#1]{\textcolor{red}{\ref*{#1}}}%
  }%
} 

\SetKwComment{rtab}{}{} 
\newcommand{\setassignlhs}[1]{\settowidth{\AssignLHS}{#1}}

\newcommand{\assignline}[3][]{%
  \begingroup
    \dimen0=\dimexpr\AssignLHS + 2\AssignGap + \ArrowW\relax
    \noindent\hangindent=\dimen0 \hangafter=1
    \rightskip=\EqTab
    \makebox[\AssignLHS][l]{#2}%
    \hspace{\AssignGap}$\leftarrow$\hspace{\AssignGap}%
    #3%
    \if\relax\detokenize{#1}\relax\relax\else\eqmark{#1}\fi
    \par\vspace{-0.2ex}%
  \endgroup
}

\newcommand{\assignlinea}[3][]{%
  \begingroup
    \dimen0=\dimexpr\AssignLHS + 2\AssignGap + \ArrowW\relax
    \noindent\hangindent=\dimen0 \hangafter=1
    \rightskip=\EqTab
    \makebox[1.5em][l]{#2}%
    \hspace{\AssignGap}$\leftarrow$\hspace{\AssignGap}%
    #3%
    \if\relax\detokenize{#1}\relax\relax\else\eqmark{#1}\fi
    \par\vspace{-0.2ex}%
  \endgroup
}

\makeatletter
\newcommand{\startsuppcontent}{%
  \let\orig@addcontentsline\addcontentsline
  \renewcommand{\addcontentsline}[3]{%
    \orig@addcontentsline{##1}{##2}{##3}%
    \addtocontents{supp}{\protect\contentsline{##1}{##2}{##3}}%
  }%
}

\newcommand{\stopsuppcontent}{%
  \let\addcontentsline\orig@addcontentsline
}

\newcommand{\supplementtoc}{%
  \section*{Supplementary Table of Contents}%
  \@starttoc{supp}%
}
\makeatother

\newcommand{\SuppTOCLine}[2]{%
  \noindent\makebox[\linewidth][l]{#1\dotfill\ #2}%
}

%% file: sec/0_abstract.tex
\begin{abstract}

Diffusion models (DMs) produce high-quality images, yet their sampling remains costly when adapted to new domains. Distilled DMs are faster but typically remain confined within their teacher's domain. Thus, fast and high-quality generation for novel domains relies on two-stage pipelines: Adapt-then-Distill or Distill-then-Adapt. However, both add design complexity and often degrade quality or diversity.
We introduce Uni-DAD, a single-stage pipeline that unifies DM distillation and adaptation. It couples two training signals: (i) a dual-domain distribution-matching distillation (DMD) objective that guides the student toward the distributions of the source teacher and a target teacher, and (ii) a multi-head generative adversarial network (GAN) loss that encourages target realism across multiple feature scales. The source domain distillation preserves diverse source knowledge, while the multi-head GAN stabilizes training and reduces overfitting, especially in few-shot regimes. The inclusion of a target teacher facilitates adaptation to more structurally distant domains.
We evaluate Uni-DAD on two comprehensive benchmarks for few-shot image generation (FSIG) and subject-driven personalization (SDP) using diffusion backbones.
It delivers better or comparable quality to state-of-the-art (SoTA) adaptation methods even with less than 4 sampling steps, and often surpasses two-stage pipelines in quality and diversity\footnote{Code: \url{https://github.com/yaramohamadi/uni-DAD}}.

\end{abstract}

%% file: sec/1_intro.tex
\section{Introduction}
\label{sec:intro}

\begin{figure}[t] 
  \centering
  \includegraphics[width=\linewidth]{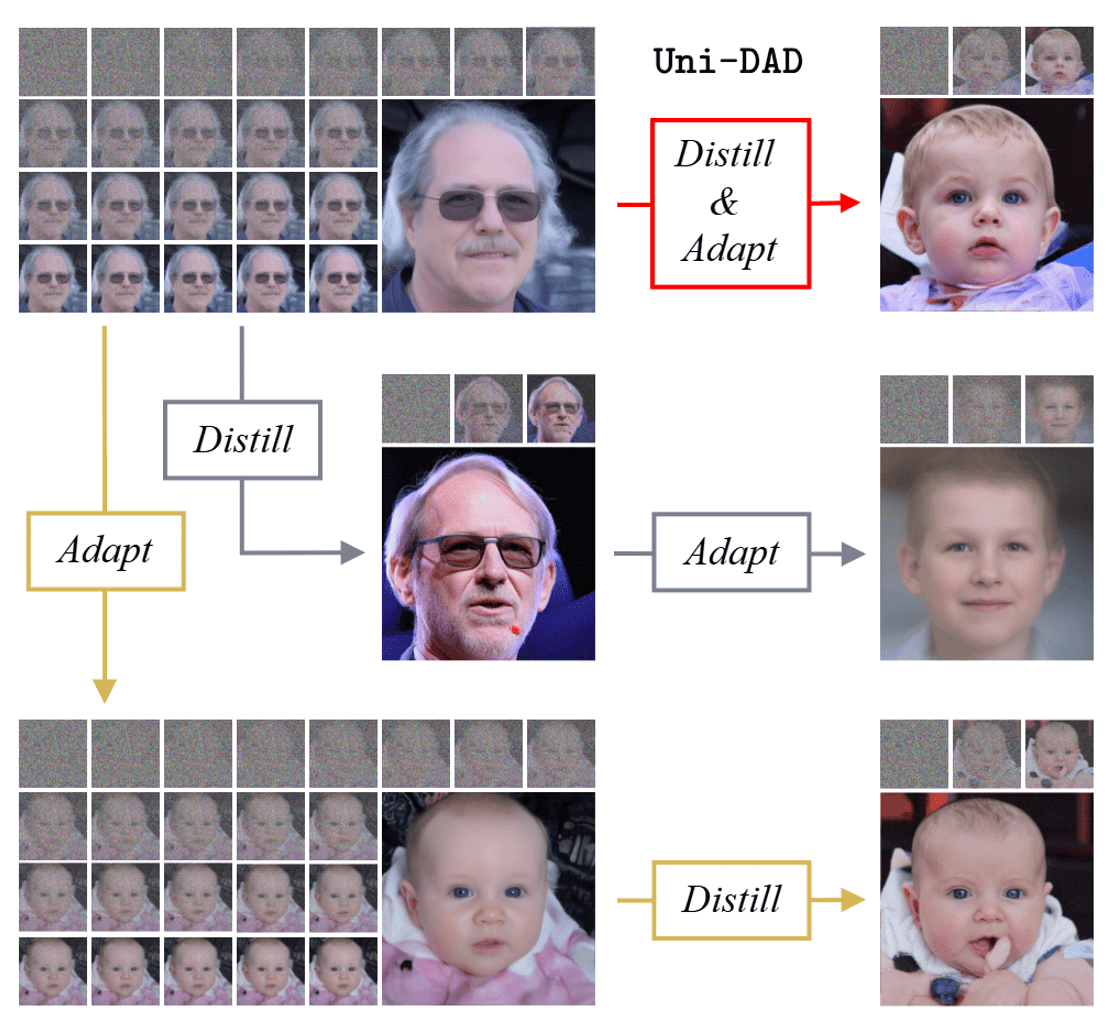}
  \caption{\outlinebox{~\texttt{Uni-DAD}~} (\emph{Distill \& Adapt}) vs. two-stage pipelines,\outlineboxx{\emph{~Distill-then-Adapt~}}, and\outlineboxxx{\emph{~Adapt-then-Distill~}}.
\textit{Adapt} is performed by fine-tuning, and \textit{Distill} by DMD2~\cite{yin2024improved}. The source domain is represented by 70K diverse faces, and the target domain by 10 babies. Sampling steps are reduced from 25 to 3.}
\label{fig:1_motivating}
\vspace{-8px}
\end{figure}





DMs~\citep{sohl2015deep,ho2020denoising,song2020score} have emerged as the dominant paradigm for generative modeling, achieving SoTA performance in image synthesis~\citep{dhariwal2021diffusion} and text-to-image generation~\citep{rombach2022high, saharia2022imagen}. These models produce high-quality and diverse images even when adapted to novel domains and subjects, given only a handful of images. This makes them an attractive solution for FSIG~\citep{cao2024few,zhu2022few} and subject-driven personalization (SDP)~\citep{gal2022textual,ruiz2023dreambooth,kumari2023custom}. However, DMs need an iterative denoising procedure over many time-steps for sampling, resulting in slow test-time generation. Adapted models inherit this cost, challenging real-time personalized use-cases. Distillation alleviates the slow inference by training a few-step student to mimic a larger teacher DM~\citep{salimans2022progressive,song2023consistency,yin2024one,yin2024improved,chadebec2025flash}. 
Ultimately, the ability to generate images in novel domains in few-shot contexts while requiring only a few denoising steps can facilitate the deployment of DMs in real-time personalized applications.





In recent works, reducing the number of time-steps and adapting to new domains requires a two-stage pipeline: \emph{Distill-then-Adapt} or \emph{Adapt-then-Distill}. The former is more compute-friendly as a student can be adapted per task after a single compute-heavy distillation step. Yet, students often saturate their adaptation capacity, yielding over-smoothed outputs on few-shot target domains~\citep{miao2024tuning}. Further, fine-tuning a student under the teacher’s original diffusion loss negates the benefits of distillation~\citep{miao2024tuning}. \emph{Adapt-then-Distill} can yield higher image quality and mitigate over-smoothing. However, the student remains tied to the adapted teacher's performance and is prone to overfitting. Conversely, neither of the two-stage pipelines is an end-to-end process, and both are susceptible to losing diverse transferable source information during the training processes.  

We propose \underline{Uni}fied \underline{D}istillation and \underline{A}daptation of \underline{D}iffusion models (\texttt{{Uni-DAD}}), a single-stage pipeline that compresses a high-quality DM teacher into a few-step student, while adapting it to a few-shot target domain (Fig.~\ref{fig:1_motivating}). It couples two complementary signals: a dual-domain DMD objective and a multi-head GAN loss. The dual-domain DMD guides the student's generation with the scores of a frozen source-domain teacher and optionally an online target-domain teacher. The inclusion of the target teacher improves adaptation to structurally distant domains. This dual-domain design guides the student toward a common area between the two distributions while preserving source diversity. The multi-head GAN enforces target realism across multiple feature scales, reducing overfitting in few-shot regimes. An online fake teacher tracks the evolving student distribution and provides up-to-date negatives for the discriminator in the GAN framework. \texttt{Uni-DAD} training iterates among updating (i) the student, (ii) the online fake teacher and discriminator, and optionally (iii) an online target teacher. These objectives allow the student to preserve source-derived diversity while sharpening target realism. The end result is a few-step generator that produces diverse high-quality images of a novel domain in few-shot contexts. \texttt{Uni-DAD} is checkpoint-agnostic: a pre-adapted target DM can replace the online target teacher with no additional training needed, and a pre-distilled source DM can initialize the student. As a result, our method enables distillation of adapted models and adaptation of distilled models without any changes to the training loop.

\texttt{Uni-DAD} is extensively validated on two benchmarks 
across different datasets and diffusion backbones: FSIG~\cite{ojha2021few} with guided denoising diffusion probabilistic model (DDPM)~\cite{dhariwal2021diffusion}, SDP~\cite{ruiz2023dreambooth} with Stable Diffusion (SD-v1.5)~\cite{luo2023latent}.
It attains better or comparable quality than SoTA adaptation methods while requiring substantially fewer sampling steps ($\leq4$), and often outperforms two-stage pipelines in quality and diversity. Our pipeline offers a practical path to fast, personalized image generation.


\noindent\textbf{Contributions.} 
\textbf{(i)}  We introduce the first single-stage pipeline that jointly distills and adapts a DM for fast, high-quality, and diverse generation in novel domains.
\textbf{(ii)} Dual-domain DMD and multi-head GAN losses are proposed to help retain source-domain diversity while sharpening target domain realism under few-shot data. An optional target teacher facilitates adaptation to structurally distant domains.
\textbf{(iii)} On FSIG and SDP benchmarks, our method achieves higher or comparable quality with substantially fewer steps than prior non-distilled adaptation methods and often outperforms two-stage pipelines in quality and diversity.






%% file: sec/2_related.tex
\section{Related Work}
\label{sec:related}

\noindent\textbf{(a) Diffusion Distillation.}
\label{sec:diffusion_distillation}
To address the costly DM inference, efficient numerical solvers (e.g., DDIM~\citep{song2020denoising}, DPM-Solver~\citep{lu2022dpm}) compress the long denoising trajectory without retraining the DM $\big(\text{neural function evaluations, NFE}\!\sim\!10\big)$. Knowledge distillation, on the other hand, trains a few-step student generator to mimic a teacher DM ($1\leq\text{NFE}\!\leq\!4$). Progressive distillation halves steps by matching the teacher across adjacent time-steps~\citep{salimans2022progressive}. Consistency-based methods directly learn a one-step mapping from noise to data~\citep{song2023consistency,luo2023latent}. Recently, by combining score distillation with adversarial training, DMD~\citep{yin2024one,yin2024improved}, ADD~\citep{sauer2024adversarial,sauer2024fast}, and FlashDiffusion~\citep{chadebec2025flash}, yield few-step students that match or surpass their teacher in quality. However, the students remain tied to the teacher’s manifold, limiting flexibility under domain shift. Furthermore, the adversarial objective assumes access to large training corpora that is unavailable in few-shot applications, 
where the GAN's discriminator easily memorizes the target set. We focus on distillation in the face of domain-shift and few-shot target sets.

\begin{figure*}[t]  
  \centering
  \includegraphics[width=\textwidth]{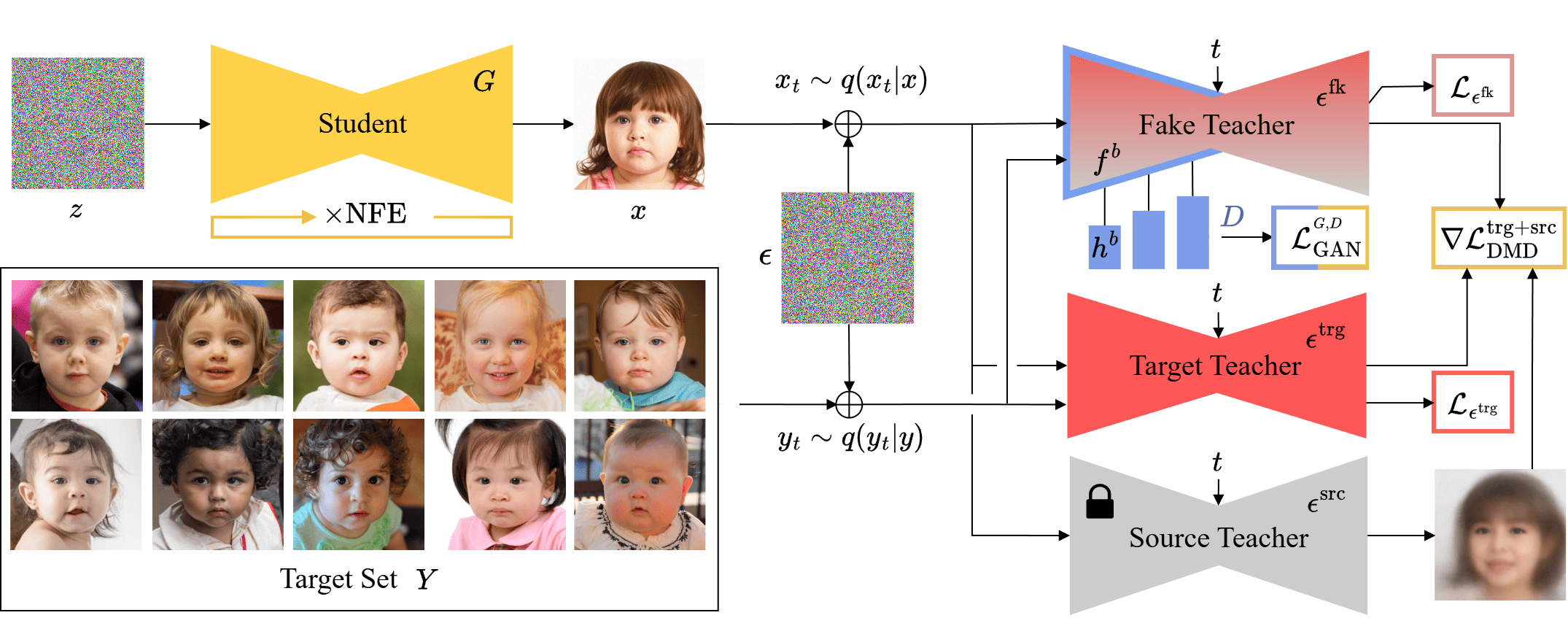}  
  \caption{
  Overview of \texttt{Uni-DAD} for few-step and few-shot image generation. A (frozen) \tboxtext{CCCCCC}{\strut source teacher $\epsilon^{\text{src}}$} is adapted and distilled into a \tboxtext{FFD24A}{\strut student $G$} for fast sampling ($1\leq\text{NFEs}\leq4$) on the target domain. At each training iteration, \texttt{Uni-DAD} alternates among three updates: \textbf{(1) Student:} optimize $G$ with a dual-domain DMD objective on $\epsilon^{\text{src}}$ and \tboxtext{FF5656}{\strut target teacher $\epsilon^{\text{trg}}$},
  plus a GAN generator loss; \textbf{(2) Fake teacher and discriminator:} train a \tboxtext{DD9692}{\strut fake teacher {$\epsilon^{\text{fk}}$}} on student generations and train a \tboxtext{7C9DE8}{\strut multi-head discriminator $D$} to distinguish target images from student generations; \textbf{(3) Target teacher update:} train $\epsilon^{\text{trg}}$ on target images.
}
  \label{fig:3_main_method}
\vspace{-8px}
\end{figure*}

\noindent\textbf{(b) Diffusion Adaptation.}
\label{sec:diffusion_adaptation}
Adaptation involves updating a model pretrained on a large source domain to fit a related, smaller target domain. While naïve finetuning is standard for style transfer with ample data~\((\text{target size } n \sim 1000)\)~\cite{hu2021lora}, it easily leads to overfitting and diversity degradation in few-shot regimes ($n\!\le\!10$). This has motivated methods tailored to few-shot applications that preserve source-domain diversity~\cite{ruiz2023dreambooth, cao2024few}. \textbf{FSIG} aims to synthesize diverse, high-quality samples from an unconditional target domain. Early progress included GAN-based approaches like cross-domain correspondence (CDC)~\citep{ojha2021few}, RiCK~\citep{zhao2023exploring}, and GenDA~\citep{yang2023one}. Diffusion-based FSIG has become prominent for its quality: pairwise adaptation (DDPM-PA) applies CDC-style regularization~\citep{zhu2022few} and conditional diffusion relaxing inversion (CRDI) learns a sample-wise guidance without base-model fine-tuning~\citep{cao2024few}. In both cases, however, sampling remains slow, leaving diffusion-based FSIG substantially slower than GANs. 
The goal of \textbf{SDP} is to adapt a DM to synthesize personalized images of a subject in novel textual contexts while preserving its identity. Textual Inversion~\cite{gal2022textual} learns a subject embedding tied to a rare token, whereas DreamBooth~\cite{ruiz2023dreambooth} fine-tunes the DM on the subject images with a unique identifier. 
Subsequent works emphasize improving training efficiency~\citep{han2023svdiff, wei2023elite}
, but sampling still remains slow. 
This paper instead focuses on fast sampling while preserving diversity and generation quality in both applications.
\noindent\textbf{(c) Combining Adaptation and Distillation.}
Transforming a large source-domain model to a smaller target domain one is commonly staged as \emph{Adapt-then-Distill}, \emph{Distill-then-Adapt}, or \emph{Distill and Adapt}~\citep{yao2021adapt}. Below, we provide an analysis for each pipeline through the lens of DMs. 

\noindent {\emph{- Distill-then-Adapt:}} A one-time distillation followed by downstream adaptation is attractive for efficiency, as distillation is typically more costly than adaptation. However, prior work suggests that students often saturate in adaptation capacity. Specifically, na\"{i}vely fine-tuning the student with the original diffusion loss negates the benefits of distillation, yielding blurry, low-detail samples~\citep{miao2024tuning} (Fig.~\ref{fig:1_motivating}, middle). PSO~\citep{miao2024tuning} trains the student with a relative likelihood objective, alleviating the blur under lightweight style transfer in larger data regimes~($n\!\sim\!1000$). However, output remains over-smoothed in few-shot contexts such as SDP. Several distillation methods emphasize producing students that remain \textsc{LoRA}-friendly after distillation~\citep{luo2023lcm,yin2024improved,chadebec2025flash}. However, LoRA-adapted models lag full finetuning under stronger distribution shifts and few-shot data. 

\noindent \emph{- Adapt-then-Distill:} Adapting the teacher to the target domain before distillation can mitigate the over-smoothed generations observed in \emph{Distill-then-Adapt} pipelines. Moreover, a GAN objective during the distillation can potentially alleviate source-domain leakage and inconsistent fitting of a fine-tuned teacher~\citep{wang2024bridging} (Tab.~\ref{fig:4_FSIG_qualitative}). However, distillation on few-shot target data remains highly prone to overfitting~(Sec.~\ref{sec:diffusion_distillation}-b). Further, the student loses access to transferable source information and its generation quality remains tied to the adapted teacher, inheriting any mis-adaptation.

\noindent {\emph{- Distill and Adapt:}} To our knowledge, no prior work performs single-stage distillation and adaptation of DMs.  Codi~\citep{mei2024codi} comes close to our task. It jointly adapts an unconditional teacher to image-conditioned tasks (inpainting and super-resolution) while distilling it to a few-step student. Its focus, however, is on providing controllability within the teacher manifold rather than few-shot adaptation to off-manifold target domains. A complementary line distills and adapts classifier-free guidance~\cite{ho2022classifier}: Plug-and-play guidance distillation~\cite{hsiao2024plug} learns a modular guidance head that can be connected to adapted models, while DogFit~\cite{bahram2025dogfit} integrates guidance distillation into transfer learning. These approaches cut NFE in half by removing the two-step cost of guidance, but do not reduce the \emph{number of denoising steps}, which is the focus of this paper.

%% file: sec/3_method.tex
\section{Proposed Method}
\label{sec:method}




\subsection{Background on DMs}
DDPMs~\citep{ho2020denoising, song2020score, sohl2015deep} are generative models that learn to reverse a fixed noising process applied over $T$ time-steps. Starting from a clean image $x$, noise  $\epsilon$ is gradually added to produce a sequence of noisy images $\{x_t\}_{t=1}^T$ where
$x_t\sim q(x_t \mid x) = \mathcal{N}\!\left(\alpha_t x,\; \sigma_t^2 I\right)$ with $\alpha_t$ and $\sigma_t$ controlling the noise schedule. A neural network $\epsilon(x_t, t)$ with parameters $\pi$ is trained to predict $\epsilon$ at each $t$, using a mean squared error (MSE) objective:
\begin{equation}
   \mathcal{L}(\pi) = \mathbb{E}_{t, x, \epsilon} \Big[ \omega_t\big\| {\epsilon}_{\pi}(x_t, t) - {\epsilon} \big\|^2 \Big],
   \label{eq:diffusion_loss}
\end{equation}
where $\omega_t>0$ is determined by the noise schedule. In subsequent equations, $t$ and $\pi$ are omitted when clear from context. In conditional generation, the model receives an auxiliary input $c$ (e.g., class labels or text prompts), producing $\epsilon(x_t|c)$. Modern DMs such as 
SDv1.5~\citep{rombach2022high, podell2023sdxl} operate in the latent rather than pixel space.



\begin{figure}[t] 
  \centering
  \includegraphics[width=\linewidth]{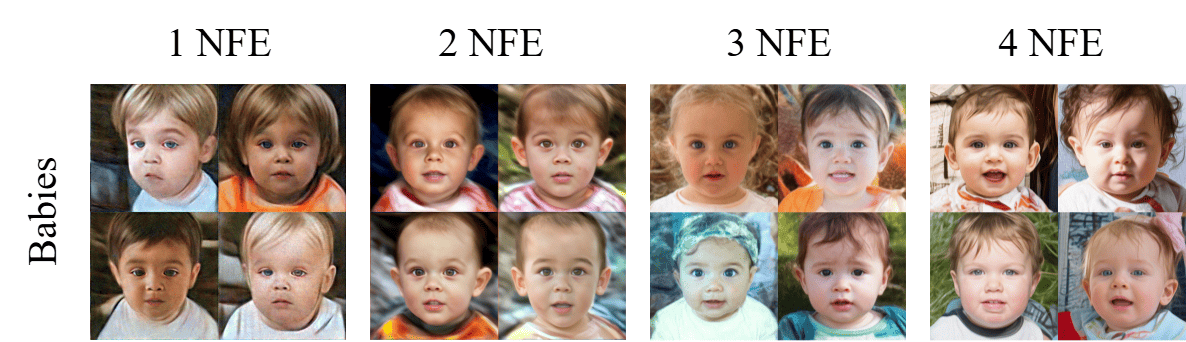}  
  \caption{Sensitivity analysis of sample quality to NFE. See Tab.~\ref{tab:4_kshot_fid} for quantitative analysis on NFE and target set size.}
  \label{fig:4_denoising_steps}
  \vspace{-8pt}
\end{figure}

\subsection{Unified Adaptation and Distillation of DMs}
\label{sec:jad}

\texttt{Uni-DAD} is proposed to compress a frozen source teacher DM $\epsilon^{\text{src}}$ trained with many time-steps ($T\sim1000$) on a large source distribution $p^{\text{src}}(x)$ into a fast student generator $G$ with parameters $\theta$ ($1\!\le\!\text{NFE}\!\le\!4$) while adapting to a target distribution $p^{\text{trg}}(y)$ represented by a few-shot target set $Y$ ($|Y|\leq10$). It couples two complementary signals for training $G$: (i) a dual-domain DMD against $\epsilon^{\text{src}}$ and optionally an online target teacher $\epsilon^{\text{trg}}$, plus (ii) a multi-head GAN loss encouraging target realism across multiple feature scales. A fake teacher $\epsilon^{\text{fk}}$ is maintained to track the evolving student distribution, with a multi-head discriminator $D$ attached to it to distinguish the student generations from $Y$. Additionally, a $\epsilon^{\text{trg}}$ can be fine-tuned on $Y$ to further improve in matching the target distribution. The training alternates among optimizing the three models, $G$, $\epsilon^{\text{fk}}+D$, and optionally $\epsilon^{\text{trg}}$ (Fig.~\ref{fig:3_main_method}). 

The next subsections detail each component: dual-domain DMD (Sec.~\ref{sec:3_weighted_dmd}), fake and target teachers (Sec.~\ref{sec:3_target_fake_teacher}), and multi-head GAN (Sec.~\ref{sec:3_multihead_gan}), while Sec.~\ref{sec:3_training_objective} describes the integration of components and overall training objective.  

\subsection{Dual-domain DMD}
\label{sec:3_weighted_dmd}

DMD is originally used to align a student's distribution $p^{\text{fk}}$ to $p^{\text{src}}$ within the source domain~\citep{yin2024one,yin2024improved}. It minimizes KL-divergence of the two distributions at the current student outputs, nudging the student generator toward higher density regions of $p^{\text{src}}$. Computing the probability densities to estimate the loss $\mathcal{L}_{\text{DMD}}(\theta)$ is generally intractable~\citep{yin2024one}. However, the gradient of this loss with respect to $\theta$ can be obtained:
\begin{align}
\label{eq:gradient_dmd}
\nabla_{\theta}\mathcal{L}_{\text{DMD}}& =\nabla_\theta D_{\mathrm{KL}}(p^{\text{fk}}\|p^{\text{src}})\; 
    \\& =\;\mathbb{E}_{\substack{z}}
\left[
\big( 
\nabla_x \log p^{\text{fk}}(x)
- 
\nabla_x \log p^{\text{src}}(x)
    \big)
    \frac{dG_{{\theta}}}{d\theta}
\right], \nonumber
\end{align}
where $x = G(z ),~z \!\sim\! \mathcal{N}(0,I)$ denotes the student output. Under Gaussian perturbation, the score satisfies
$s(x_t) = \nabla_{x_t}\log p(x_t) = -\frac{1}{\sigma_t}\,\epsilon(x_t)
$~\cite{song2020score}. Therefore, the right-hand terms of Eq.~\ref{eq:gradient_dmd} can be approximated via two DMs:  $\epsilon^{src}$, and $\epsilon^{\text{fk}}$, where $\epsilon^{src}$ is frozen and $\epsilon^{\text{fk}}$ is concurrently trained to track the evolving student outputs (Sec. ~\ref{sec:3_target_fake_teacher}). In practice, $\mathcal{L}_{\text{DMD}}$ can be minimized by updating  $\theta \leftarrow \theta - \eta \nabla_{\theta}\mathcal{L}_{\mathrm{DMD}}$ with gradient descent.

\begin{figure}[t]  
  \centering
  \includegraphics[width=\linewidth]{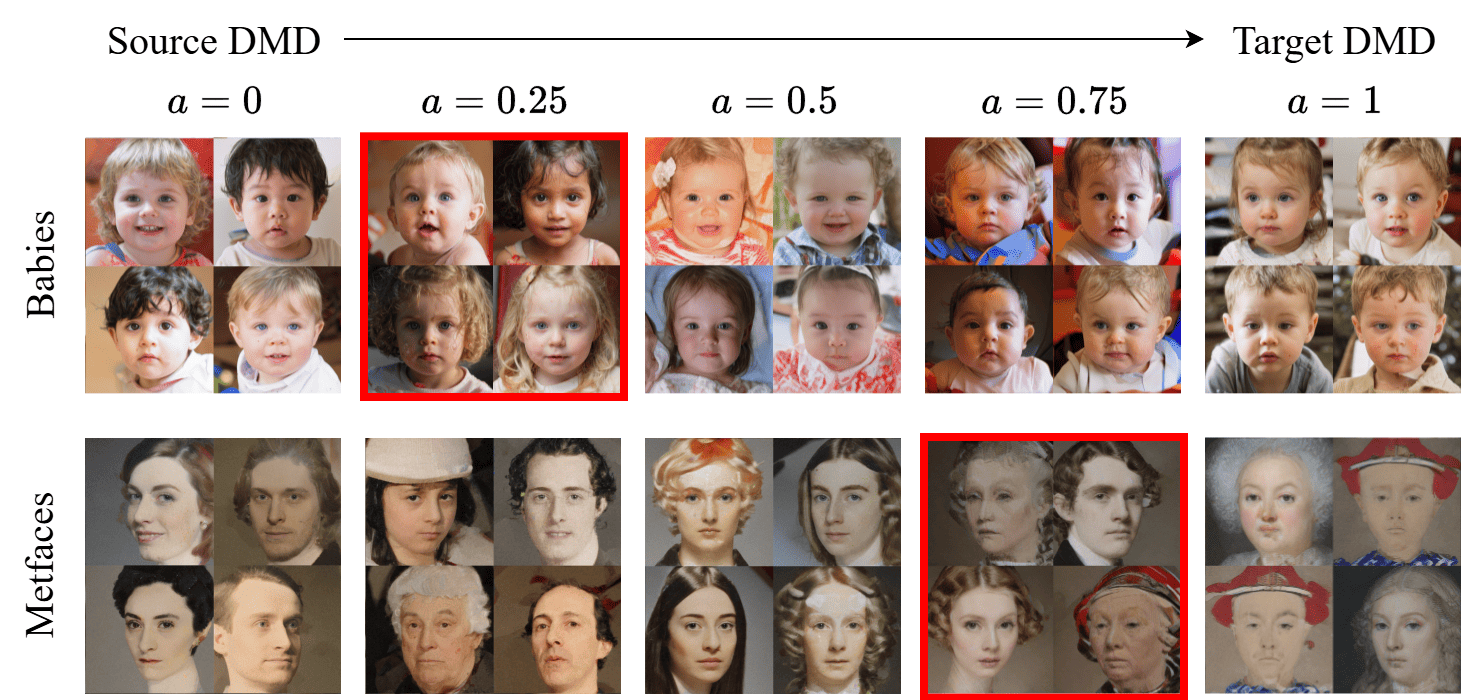} 
  \caption{Qualitative ablation of the dual-domain DMD weighting factor $a$ for FSIG on Babies and MetFaces. See Fig.~\ref{fig:SDP_alpha} for SDP.}
  \label{fig:4_ablation_a}
  \vspace{-8pt}
\end{figure}

We extend this formula to align the student outputs to both $p^{\text{src}}$ and $p^{\text{trg}}$. The gradients of the two DMD losses can be written in noise-estimation form:
\begin{align}
\label{eq:dmd_grad}
\nabla_{{\theta}} & \, \mathcal{L}_{\text{DMD}^{\text{src}}}
\;\approx\;
\mathbb{E}_{\substack{t,z}}
\left[
\omega_t \left(
\epsilon^{\text{fk}}\left(x_t \right)\;-\; \epsilon^{\textbf{src}}(x_t )
\right)
\frac{dG_{{\theta}}}{d\theta}
\right], \nonumber \\
\nabla_{{\theta}} & \, \mathcal{L}_{\text{DMD}^{\text{trg}}}
\;\approx\;
\mathbb{E}_{\substack{t,z}}
\left[
\omega_t \big(
\epsilon^{\text{fk}}(x_t )\;-\; \epsilon^{\textbf{trg}}(x_t )
\big)
\frac{dG_{{\theta}}}{d\theta}
\right],
\end{align}
where
$
t \sim \mathcal{U}\{0.02T, 0.98T\}
$
 and extreme time-steps are excluded for numerical stability~\citep{poole2022dreamfusion}. We use a normalization that balances contributions across time-steps:
\begin{equation}
\omega_t
\;=\;
\frac{\sigma_t \cdot H \cdot S}
{\bigl\| \epsilon - \epsilon^{\text{fk}}(x_t ) \bigr\|_1},
\label{eq:dmd-weight}
\end{equation}
\noindent with $H$ channels and $S$ spatial locations~\citep{yin2024one}.  Optimizing $\mathcal{L}_{\text{DMD}}^{\text{src}}$ can help retain diverse transferable information (e.g., pose, background, and facial expression), thereby compensating for target data scarcity. This objective suffices for adaptation in the face of small domain shifts. However, more structurally dissimilar target domains may contain regions outside the source manifold, in which case $\mathcal{L}_{\text{DMD}}^{\text{src}}$ can hold back true adaptation. A dual-domain DMD objective can guide the student toward a common area between the two distributions:
\begin{equation}
\nabla_{\theta}\mathcal{L}^{\text{trg}+\text{src}}_\text{DMD}=(1-a)\nabla_{\theta}\mathcal{L}_{\text{DMD}^{\text{src}}} + a\nabla_{\theta}\mathcal{L}_{\text{DMD}^{\text{trg}}},
   \label{eq:dual_dmd_grad}
\end{equation}
where $a\in[0,1]$ indicates a weighting factor, controlling the influence of each domain (see Fig.~\ref{fig:4_ablation_a} and Fig.~\ref{fig:SDP_alpha}).



\subsection{Fake and Target Teachers}
\label{sec:3_target_fake_teacher}
We initialize $\epsilon^\text{fk}$ with the weights of $\epsilon^\text{src}$ and update its parameters $\phi$ on the evolving student outputs by minimizing the MSE objective:
\begin{equation}
\mathcal{L}_{\text{fk}}({\phi})
=\mathbb{E}_{t,z}\Big[\,\big\|\epsilon{^\text{fk}}_{{\phi}}(x_t)-\epsilon\big\|_2^2\Big].
\label{eq:fake_loss}
\end{equation}

During $\epsilon^\text{fk}$ updates, no gradients are propagated through $G$, and $x$ is treated as fixed. Similarly, $\epsilon^{\text{trg}}$ is initialized with the weights of $\epsilon^{src}$ and its parameters $\eta$ are updated via the MSE to denoise diffused samples from $Y$ via:
\begin{equation}
\mathcal{L}_{\text{trg}}({\eta})
=\mathbb{E}_{t,\epsilon,y}\Big[\,\big\|\epsilon^\text{trg}_{{\eta}}(y_t)-\epsilon\big\|_2^2\Big].
\label{eq:target_loss}
\end{equation}

The training and inclusion of $\epsilon^{\text{trg}}$ is optional, as it can facilitate adaptation of structure in face of strong domain shifts (see component ablations in Tab.~\ref{tab:4_component_ablation}).
Further, if a pre-adapted $\epsilon^{\text{trg}}$ checkpoint is already available, it can be used as a fixed target teacher without further training (see Tab.~\ref{tab:4_init_ablation}).


 \begin{figure*}[!t]
  \centering
  \includegraphics[width=\textwidth]{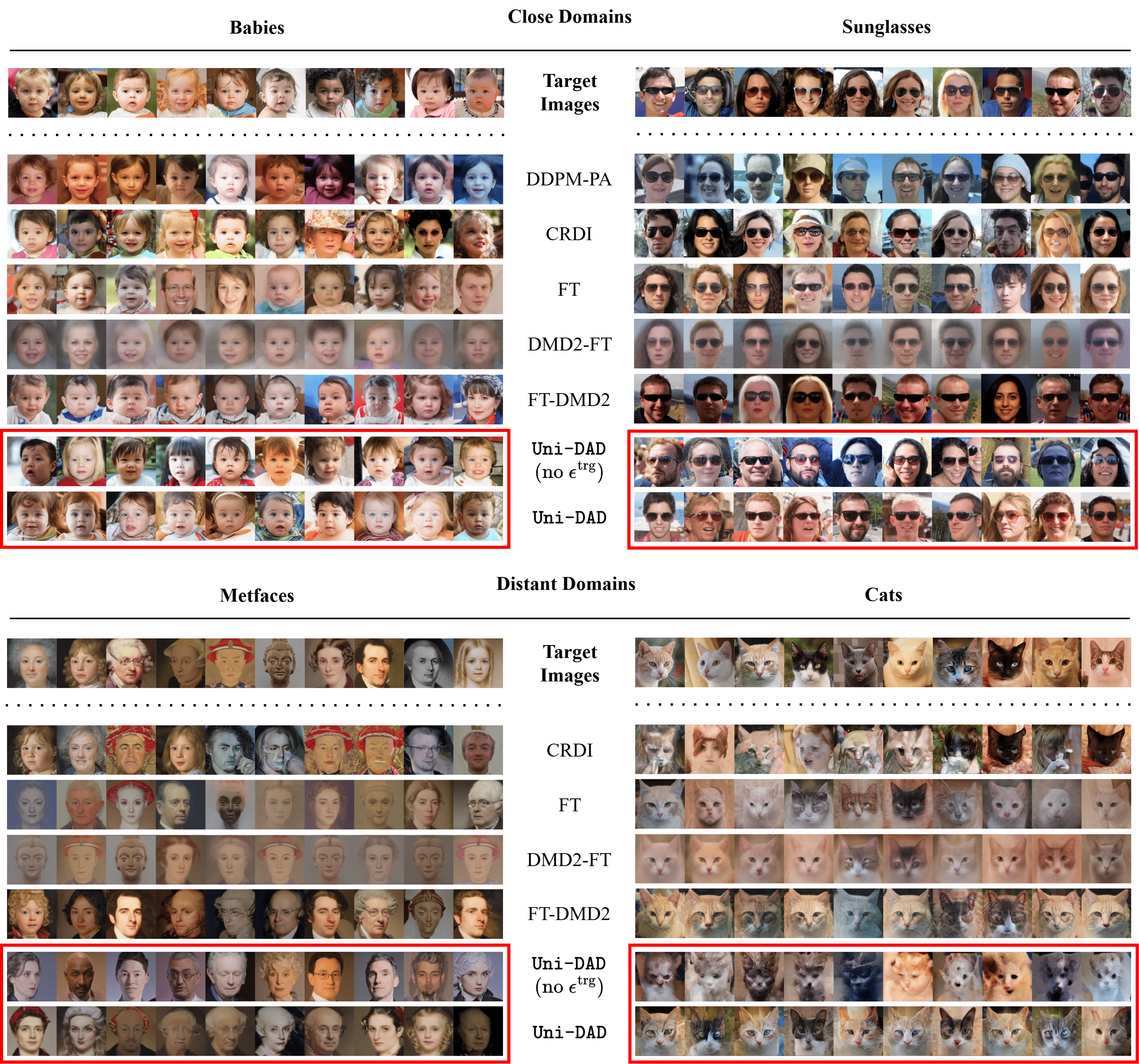}
  \caption{Qualitative comparison for FSIG, adapting guided-DDPM~\cite{dhariwal2021diffusion} pretrained on FFHQ~\cite{karras2019style} to 10-shot target sets of varying proximity to the source domain. See Fig. \ref{fig:A_all_qualitative_source} and \ref{fig:A_all_qualitative_source_target} for additional generations. Generated samples are randomly picked. Zoom in for details.}
  \label{fig:4_FSIG_qualitative}
  \vspace{-8pt}
\end{figure*}

\subsection{Multi-head GAN}
\label{sec:3_multihead_gan}

To enforce sharp fidelity of student outputs to $Y$ and stabilize training, we use a multi-head GAN objective that judges target realism at multiple feature levels. While $G$ plays the role of generator, the discriminator $D$ reuses $\epsilon^{\text{fk}}$ encoder and middle blocks for feature extraction: let $f^b(\cdot)$ denote the feature extractor at block $b \in \mathcal{B}$ of $\epsilon^{\text{fk}}$, and attach a linear head $h^l(\cdot)$ with parameters $\psi$ to every block's output. This yields a multi-head discriminator whose output at block $b$ is
$D^b(\cdot)=\sigma\!\big(h^b(f^b(\cdot))\big)$,
where $\sigma(\cdot)$ denotes the sigmoid activation. Its aim is to distinguish between $y\in Y$ and $x\!=\!G(z),~z\sim \mathcal{N}(0,I)$. The GAN losses, aggregated over heads by summation, are:
\begin{align}
\mathcal{L}_{\text{GAN}}^{G}(\theta)
&= -\,\mathbb{E}_{t,z}\!
   \sum_{b\in\mathcal{B}}
   \log\!\big(D_{\theta}^b(x_t)\big),
\label{eq:gan_g_loss}\\[4pt]
\mathcal{L}_{\text{GAN}}^{D}(\psi,\phi)
&= -\,\mathbb{E}_{t,y}\!
   \sum_{b\in\mathcal{B}}
   \log\!\big(D_{\psi,\phi}^b(y_t)\big)
\notag\\
&\quad-
   \mathbb{E}_{t,z}\!
   \sum_{b\in\mathcal{B}}
   \log\!\big(1 - D_{\psi,\phi}^b(x_t)\big).
\label{eq:gan_d_loss}
\end{align}

Attaching classifier heads after every encoder block enables $D$ to contrast real vs.\ fake at both local and global scales, which is especially helpful in the few-shot regime $|Y|\leq 10$, mitigating overfitting and mode collapse.

\subsection{Overall Training Objective}
\label{sec:3_training_objective}
The \textbf{student}'s training update balances source preservation and target fitting via minimizing the dual-domain DMD and GAN generator losses:
\begin{align}
    \mathcal{L}_G(\theta) =~ \mathcal{L}^{\text{trg}+\text{src}}_\text{DMD}(\theta) + \lambda_{\text{GAN}}^{G}\mathcal{L}^{G}_{\text{GAN}}(\theta),
   \label{eq:g_total_loss}
\end{align}
where in practice, $\nabla_{\theta}\mathcal{L}^{\text{trg}+\text{src}}_\text{DMD}$ is used instead of $\mathcal{L}^{\text{trg}+\text{src}}_\text{DMD}(\theta)$ to update $G$'s parameters. The \textbf{fake teacher}'s training update combines its MSE and GAN discriminator losses:
\begin{align}
     \mathcal{L}_{\text{fk}+D}(\phi, \psi) = \mathcal{L}_{\text{fk}}(\phi) + \lambda_{\text{GAN}}^{D}\mathcal{L}^{D}_{\text{GAN}}(\psi,\phi).
     \label{eq:fk_total_loss}
\end{align}

The training involves alternating among minimizing three losses at each iteration: $\mathcal{L}_\text{G}$, $\mathcal{L}_{\text{fk}+D}$ and $\mathcal{L}_{\text{trg}}$  (see Alg. \ref{alg:uni_dad}). In practice, the minimization of $\mathcal{L}_{\text{fk},D}$ is performed 5-10 times~\cite{yin2024improved} for each update of $\mathcal{L}_\text{G}$ and $\mathcal{L}_{\text{trg}}$ to allow $\epsilon^{\text{fk}}$ to keep up with the constantly changing output distribution of $G$. For SDP-specific methodology details, see Appx.~\ref{sec:A_SDP_methodology}.

%% file: sec/4_results.tex
\section{Results and Discussion}
\label{sec:experiments}


\subsection{Experimental Setup}
\label{sec:experimental_setup}

\input{src/tables/4_FSIG_FID}

\input{src/tables/A_FSIG_computational}

\noindent\textbf{FSIG Benchmark:} ~\cite{ojha2021few, zhao2023exploring} We use the guided DDPM~\cite{dhariwal2021diffusion} pre-trained on the unconditional FFHQ dataset (70K images of diverse faces~\cite{karras2019style}\footnote{FFHQ weights: \url{https://github.com/yandex-research/ddpm-segmentation}}) as the source model, and adapt it to 10 pre-selected samples of four target domains. We include two semantically close domains: Babies~\cite{ojha2021few} and Sunglasses~\cite{ojha2021few}, and two structurally distant domains: MetFaces~\cite{karras2020training} and AFHQ-Cat (Cats)~\cite{choi2020stargan}. We compare against DDPM-PA~\cite{zhu2022few} and CRDI~\cite{cao2024few}.
DDPM-PA provides no public code,
so we quote their numbers when available. To measure quality, we report FID on 5K generations against held-out target sets ($|\text{Babies}|$=2.5K , $|\text{Sunglasses}|$=2.7K, $|\text{MetFaces}|$=1.3K, $|\text{Cats}|$=5K). We assess diversity via Intra-LPIPS~\cite{ojha2021few} on 1K generations relative to the 10-shot targets. All adaptations use \(256{\times}256\) resolution. Unless specified, \texttt{Uni-DAD} uses 
$\text{NFE}=3$ and 10-shot target sets.

\noindent\textbf{SDP Benchmark:}~\cite{ruiz2023dreambooth} We use SDv1.5~\cite{rombach2022high} pretrained on LAION-5B~\cite{schuhmann2022laion} as the source model. We evaluate on the DreamBooth benchmark containing 30 subjects
each represented by 4-6 images. For each target subject, we generate 100 samples (25 text prompts $\times$ 4 seeds).
We compare against
DreamBooth~\cite{ruiz2023dreambooth} 
and PSO~\cite{miao2024tuning}. PSO adapts a Turbo-distilled model~\cite{chadebec2025flash} on an SDXL backbone~\cite{podell2023sdxl},
but provides no code for SDv1.5, so we report their results despite the resolution and backbone mismatch giving an unfair advantage to PSO. Identity preservation is measured using DINO (ViT-S/16) similarity~\cite{caron2021emerging} and CLIP-I (ViT-B/32) cosine similarity. Text–image alignment is quantified using CLIP-T (ViT-B/32)~\cite{radford2021learning}. We use Intra-LPIPS, and Inter-LPIPS for diversity assessment. All adaptations use 512×512 resolution except PSO  ($1024\times1024$). Unless specified, \texttt{Uni-DAD} uses $\text{NFE}=1$ and 4-6-shot target sets.



\noindent\textbf{Additional Baselines:} We consider a family of two-stage baselines across FSIG and SDP, adapted to each benchmark setting. For both tasks, we implement: (i) FT, where we fine-tune the source DM on the target domain; (ii) DMD2-FT, where we first distill the DM via DMD2~\cite{yin2024improved} and then fine-tune the student on the target domain; and (iii) FT-DMD2, where we first fine-tune the source DM and then distill it via DMD2. We use DreamBooth~\cite{ruiz2023dreambooth} as the fine-tuning method in SDP. 

\noindent\textbf{Training Details:} See Appx.~\ref{sec:A_details_FSIG} for FSIG and~\ref{sec:A_details_SDP} for SDP.




\subsection{FSIG Results}

\noindent \textbf{(a) Qualitative.} Fig.~\ref{fig:4_FSIG_qualitative} presents a qualitative comparison against FSIG SoTA ,methods on both close and distant target domains. \emph{Non-distilled} variants generally suffer from weak adaptation or unstable fitting: DDPM-PA exhibits reduced detail and noticeable color shifts. 
CRDI tends to remain close to the source manifold with limited attribute recombination, frequently regenerating the same target exemplars with minor local variations. FT suffers from inconsistent fitting, either leaking source-domain characteristics or overfitting to a few target exemplars. Among \emph{Distilled} variants, DMD2-FT naïvely nullifies the benefits of distillation, producing over-smoothed outputs with muted textures and limited diversity. While FT-DMD2 relatively improves fidelity, it still frequently collapses toward a small subset of target samples.
In contrast, \texttt{Uni-DAD} consistently yields sharp, high-quality, and diverse generations. It mitigates inconsistent fitting by effectively finding a middle ground between source preservation and target fitting, and better combines target attributes. Including the target teacher further improves structural adaptation on distant domains, at the cost of a slight reduction in diversity. See Appx.~\ref{sec:A_FSIG_additional} for additional generated samples.



\noindent \textbf{(b) Quantitative.}
Tab.~\ref{tab:4_fid_lpips_comparison} reports the quantitative comparison. 
Compared to \emph{Non-distilled} baselines, \texttt{Uni-DAD} consistently achieves better FID while using only 3 denoising steps, indicating higher generation quality at substantially lower sampling cost. Moreover, its Intra-LPIPS remains comparable to \emph{Non-distilled} methods despite the diversity reduction commonly observed in distilled models~\citep{gandikota2025distilling}. Compared to other \emph{Distilled} variants, \texttt{Uni-DAD} obtains stronger FID and Intra-LPIPS on \textbf{close domains}. On \textbf{distant domains}, FT-DMD2 becomes more competitive; however, by incorporating the target teacher, \texttt{Uni-DAD} can achieve similar FID, with only a slight reduction in Intra-LPIPS. Overall, these results show that \texttt{Uni-DAD} remains effective across both close and distant domain shifts.

\noindent \textbf{(c) Computational Cost.}
Tab.~\ref{tab:A_Compute_cost} compares training and test-time costs across adaptation pipelines. While \emph{Non-distilled} variants require long denoising trajectories (NFE$=25$), \emph{distilled} methods, including \texttt{Uni-DAD}, reduce generation time for 5K samples from 35--63 minutes to 4.2 minutes and lower per-image cost from 55.7 to 2.2 TFLOPs. This efficient inference comes at the price of training compute. Nevertheless, \texttt{Uni-DAD} requires lower training cost than the two-stage baselines: 2.2 GPU$\cdot$h without a target teacher, and 2.8 GPU$\cdot$h with it, versus 3 GPU$\cdot$h for the two-stage pipelines. The target teacher training can increase peak training memory to 48.8 GB (21\% more). Designing more parameter-efficient variants of \texttt{Uni-DAD} is for future work.

\input{src/tables/4_SDIG_qualitative_dog6}

\input{src/tables/4_SDIG_Quantitative_Overall}

\subsection{SDP Results}
\label{sec:A_SDP_results}



\noindent \textbf{(a) Qualitative:} Fig.~\ref{fig:4_SDP_qualitative} presents a comparison on the DreamBooth~\cite{ruiz2023dreambooth} \emph{cat2} subject across two prompts. Compared with the \emph{Non-distilled} FT, \texttt{Uni-DAD} allows comparable subject fidelity and prompt alignment while using $\times 100$ fewer NFEs. Compared with the \emph{Distilled} baselines, FT-DMD2 and DMD2-FT, it consistently produces sharper and more faithful generations while more strongly following the prompt. In contrast, DMD2-FT exhibits severe over-smoothing and loss of detail, while FT-DMD2 tends to overfit and show weak prompt alignment. \emph{Distilled} Turbo-PSO~\cite{miao2024tuning} achieves strong prompt alignment and subject fidelity, but its generations are over-smoothed. That said, it benefits from a stronger backbone, higher resolution, and larger NFE budget, and is therefore not directly comparable to our setting. Overall, \texttt{Uni-DAD} can retain personalization quality despite extreme sampling reduction. See Appx.~\ref{sec:A_SDP_additional} for additional qualitative results and generated samples.

\noindent \textbf{(b) Quantitative:} Tab.~\ref{tab:sdp_overall} shows the quantitative comparison. While operating in a strict 1-step regime, \texttt{Uni-DAD} achieves strong DINO and CLIP-based identity and text-alignment scores, remaining comparable to \emph{Non-distilled} FT and outperforming the 1-step \emph{Distilled} baselines. In particular, FT-DMD2 attains stronger DINO and CLIP-I scores but suffers from a severe drop in diversity, whereas DMD2-FT preserves diversity better but at the cost of weaker quality and identity preservation. Overall, \texttt{Uni-DAD} provides the best trade-off between subject fidelity, prompt alignment, and diversity among the distilled methods.

\subsection{Ablations}

\noindent \textbf{(a) Weighting Factor $a$.} 
Fig.~\ref{fig:4_ablation_a} ablates coefficient $a$~(Eq.~\ref{eq:gradient_dmd}). When the target domain lies close to the source manifold (e.g., Babies), a small value is sufficient, whereas larger values help adapt to more structurally dissimilar domains (e.g., MetFaces). In practice, overly small values can restrict the student to style-transfer behavior, whereas overly large values may lead to overfitting and increased sensitivity to imperfections in the target teacher. Careful selection of $a$ allows the source term to compensate for target-teacher errors while enabling adaptation to novel structures. This trade-off is reflected in our FSIG experiments (Fig.~\ref{fig:4_FSIG_qualitative} and Tab.~\ref{tab:4_fid_lpips_comparison}), where the target teacher is removed ($a=0$) under mild domain shifts. See Fig.~\ref{fig:SDP_alpha} for SDP.



\noindent \textbf{(b)-(e):} See Appx.~\ref{sec:A_ablations_FSIG} for more ablations on FSIG. 

\noindent\textbf{(f)-(g):} See Appx.~\ref{sec:A_ablations_SDP} for ablations on SDP.


%
%

%% file: src/tables/4_FSIG_FID.tex
\begin{table*}[t]
\centering\footnotesize
\caption{Comparison of FID$\downarrow$ and Intra-LPIPS$\uparrow$ for FSIG across methods and 10-shot target sets. \textbf{Bold} indicates best result among \emph{distilled} variants. \underline{Underline} indicates best result among \emph{all} models.}
\label{tab:4_fid_lpips_comparison}
\setlength{\tabcolsep}{3.5pt}
\begin{tabularx}{\textwidth}{l l c c *{8}{Y}}
\toprule
& & & & \multicolumn{4}{c}{\textbf{FID $\downarrow$}} & \multicolumn{4}{c}{\textbf{Intra-LPIPS $\uparrow$}} \\
\cmidrule(lr){5-8} \cmidrule(lr){9-12}
\textbf{Variant} & \textbf{Method} & \textbf{NFE$\downarrow$} & \textbf{1-stage} 
       & Babies & Sunglasses & MetFaces & Cats 
       & Babies & Sunglasses & MetFaces & Cats \\
\midrule

\multirow{1}{*}{\emph{Non-distilled}} & DDPM-PA~\cite{zhu2022few}     & 1000 & \checkmark & 48.92 & 34.75 & \na & \na & \underline{0.59} & \underline{0.60} & \na & \na \\
 & CRDI~\cite{cao2024few}         & 25   & \checkmark & 48.52 & 24.62 & 121.36\textsuperscript{†} & 220.95 & 0.52 & 0.50 & 0.41 & \underline{0.51} \\

 & FT                             & 25   & \checkmark & 57.06 & 37.86 & 72.99 & 61.62 & 0.32 & 0.48 & \underline{0.45} & 0.42 \vspace{0.5mm} \\
\midrule
\multirow{1}{*}{\emph{Distilled}} 
 & DMD2-FT  & \textbf{3} &         & 140.27 & 77.49 & 129.26 & 89.32 & 0.08 & 0.20 & 0.08 & 0.18 \\
 & FT-DMD2  & \textbf{3} &         & 57.11 & 41.85 & 63.25 & \textbf{\underline{51.85}} & 0.42 & 0.42 & \textbf{0.44} & 0.34 \\\vspace{0.4mm} 
 & \cellcolor{black!5}\texttt{Uni-DAD} (no $\epsilon^{\text{trg}}$)
   & \cellcolor{black!5}\textbf{3} 
   & \cellcolor{black!5}\checkmark 
   & \cellcolor{black!5} 47.38
   & \cellcolor{black!5} \textbf{\underline{22.57}}
   & \cellcolor{black!5} 72.18
   & \cellcolor{black!5} 199.91
   & \cellcolor{black!5} 0.45
   & \cellcolor{black!5} 0.51
   & \cellcolor{black!5} 0.42
   & \cellcolor{black!5} \textbf{0.40} \\
 & \cellcolor{black!5}\texttt{Uni-DAD} 
   & \cellcolor{black!5}\textbf{3} 
   & \cellcolor{black!5}\checkmark 
   & \cellcolor{black!5}\textbf{\underline{45.09}} 
   & \cellcolor{black!5}24.45
   & \cellcolor{black!5}\textbf{\underline{58.13}} 
   & \cellcolor{black!5}55.32 
   & \cellcolor{black!5}\textbf{0.46} 
   & \cellcolor{black!5}\textbf{0.54} 
   & \cellcolor{black!5}0.36 
   & \cellcolor{black!5}0.36 \\
\bottomrule
\end{tabularx}

\vspace{3pt}
\raggedright\footnotesize
† We were unable to reproduce CRDI's reported FID of 94.86~\cite{cao2024few}. However, our qualitative samples and Intra-LPIPS closely match their findings.
\end{table*}

%% file: src/tables/A_FSIG_computational.tex
\begin{table*}[t]
\centering
\footnotesize
\setlength{\tabcolsep}{4pt}
\caption{Training and test-time computational cost analysis for FSIG. Mem: memory. h: hour, m: minute.}
\label{tab:A_Compute_cost}
\begin{tabularx}{\textwidth}{l l YYYYYYY}
\toprule
& & \multicolumn{3}{c}{\textbf{Training Cost (20K iters, †: not applicable)}} 
  & \multicolumn{4}{c}{\textbf{Test Cost (5K generations, batch size 10)}} \\
\cmidrule(lr){3-5} \cmidrule(lr){6-9}
\textbf{Variant} 
& \textbf{Method} 
& $\text{GPU}\cdot\text{h}$
& GPU 
& Mem [GB] 
& Time [m] 
& Mem [GB] 
& TFLOPs/img 
& NFE \\
\midrule
\multirow{2}{*}{\emph{Non-distilled}}
& CRDI~†                
& 1    & $4$ A100 & 124.7 & 63  & 14 & 55.7 & 25 \\
& FT                    
& 0.8  & $1$ H100 & 27.6  & 35  & 14 & 55.7 & 25 \\
\midrule
\multirow{4}{*}{\emph{Distilled}}
& FT-DMD2               
& 3    & $1$ H100 & 40.4 & 4.2 & 13 & 2.2 & 3 \\
& DMD2-FT               
& 3    & $1$ H100 & 40.4 & 4.2 & 13 & 2.2 & 3 \\
& \cellcolor{black!5}\texttt{Uni-DAD} (No $\epsilon^{\text{trg}}$) 
& \cellcolor{black!5}\textbf{2.2} 
& \cellcolor{black!5}$1$ H100 
& \cellcolor{black!5}\textbf{40.3} 
& \cellcolor{black!5}4.2 
& \cellcolor{black!5}13 
& \cellcolor{black!5}2.2 
& \cellcolor{black!5}3 \\
& \cellcolor{black!5}\texttt{Uni-DAD} 
& \cellcolor{black!5}{2.8} 
& \cellcolor{black!5}$1$ H100 
& \cellcolor{black!5}48.8 
& \cellcolor{black!5}4.2 
& \cellcolor{black!5}13 
& \cellcolor{black!5}2.2 
& \cellcolor{black!5}3 \\
\bottomrule
\end{tabularx}
\vspace{-8pt}
\end{table*}

%% file: src/tables/4_SDIG_qualitative_dog6.tex
\begin{figure*}[t]
\centering\footnotesize

\setlength{\tabcolsep}{2pt}
\newcommand{\cattarget}{\includegraphics[width=0.65\colw,height=0.65\colw,keepaspectratio]{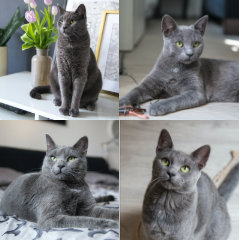}}

\newcommand{\dogtarget}{\includegraphics[width=0.65\colw,height=0.65\colw,keepaspectratio]{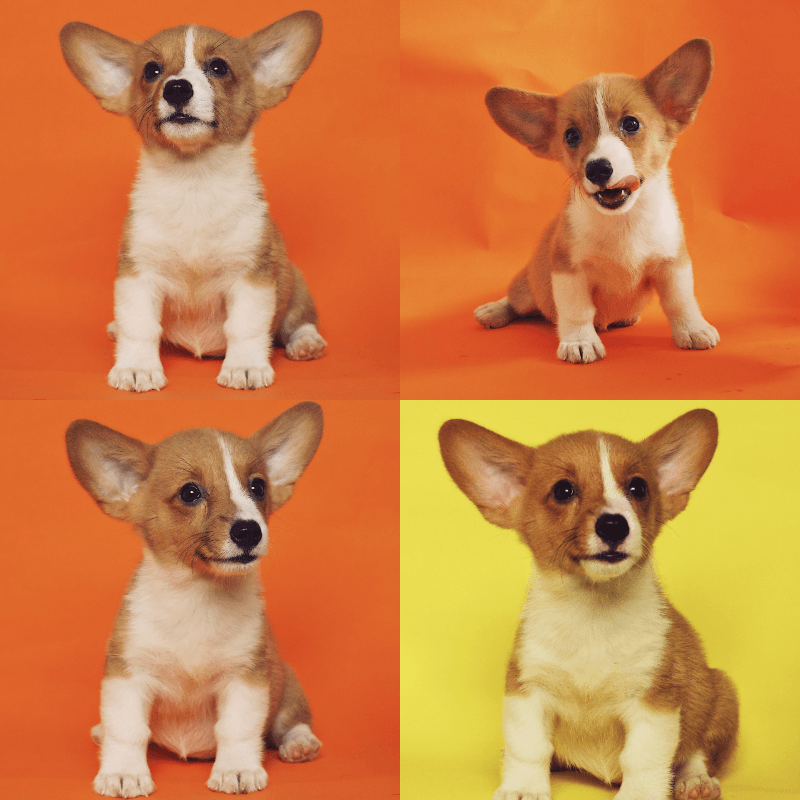}}

\newcommand{\vasetarget}{\includegraphics[width=\colw,height=\colw,keepaspectratio]{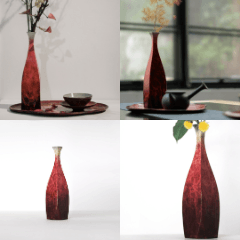}}

\newcommand{\dogunidadmountain}{\includegraphics[width=\colw,height=\colw,keepaspectratio]{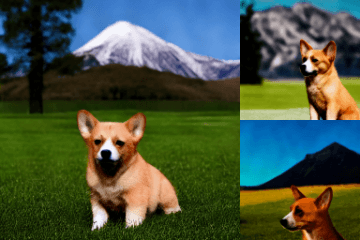}}
\newcommand{\dogunidadblue}{\includegraphics[width=\colw,height=\colw,keepaspectratio]{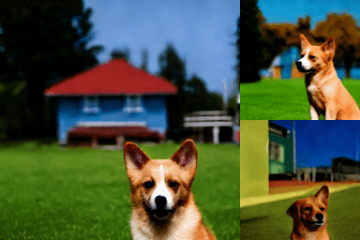}}
\newcommand{\dogunidadsnow}{\includegraphics[width=\colw,height=\colw,keepaspectratio]{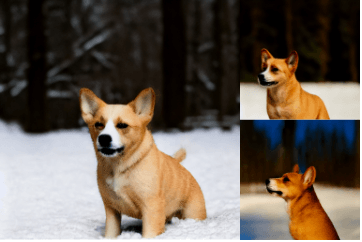}}

\newcommand{\catunidadpinkg}{\includegraphics[width=\colw,height=\colw,keepaspectratio]{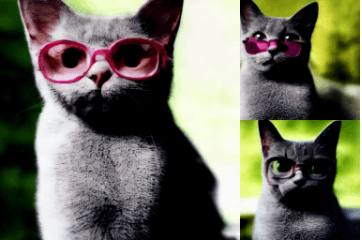}}
\newcommand{\catunidadjungle}{\includegraphics[width=\colw,height=\colw,keepaspectratio]{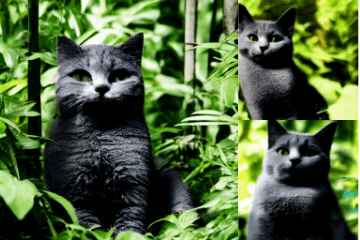}}
\newcommand{\catunidadfiref}{\includegraphics[width=\colw,height=\colw,keepaspectratio]{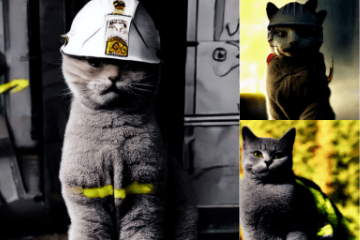}}

\newcommand{\vaseunidadbeach}{\includegraphics[width=\colw,height=\colw,keepaspectratio]{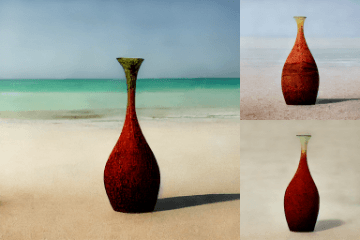}}
\newcommand{\vaseunidadsunf}{\includegraphics[width=\colw,height=\colw,keepaspectratio]{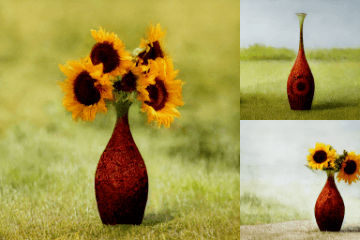}}
\newcommand{\vaseunidadcob}{\includegraphics[width=\colw,height=\colw,keepaspectratio]{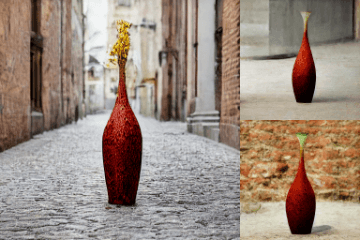}}

\newcommand{\dogdbmountain}{\includegraphics[width=\colw,height=\colw,keepaspectratio]{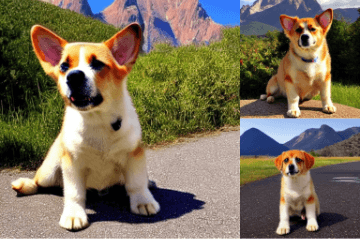}}
\newcommand{\dogdbblue}{\includegraphics[width=\colw,height=\colw,keepaspectratio]{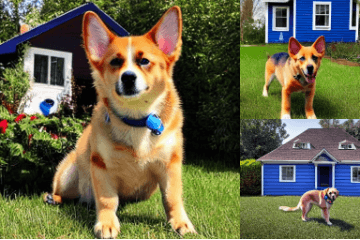}}
\newcommand{\dogdbsnow}{\includegraphics[width=\colw,height=\colw,keepaspectratio]{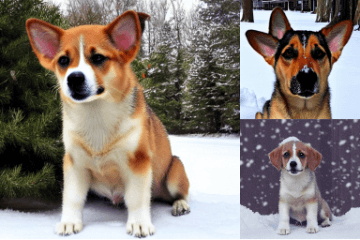}}

\newcommand{\vasedbbeach}{\includegraphics[width=\colw,height=\colw,keepaspectratio]{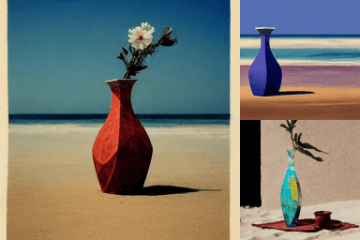}}
\newcommand{\vasedbcobb}{\includegraphics[width=\colw,height=\colw,keepaspectratio]{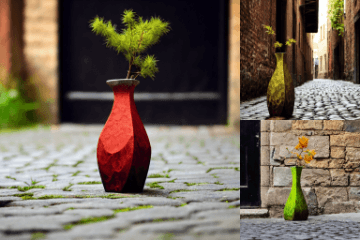}}
\newcommand{\vasedbsunf}{\includegraphics[width=\colw,height=\colw,keepaspectratio]{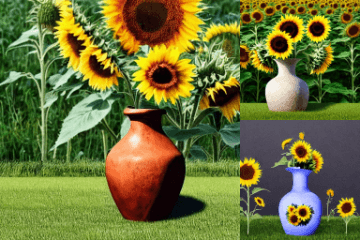}}

\newcommand{\catdbjungle}{\includegraphics[width=\colw,height=\colw,keepaspectratio]{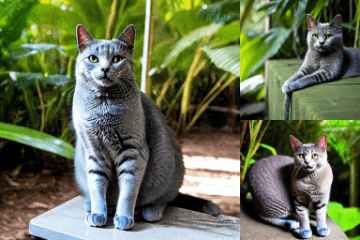}}
\newcommand{\catdbpink}{\includegraphics[width=\colw,height=\colw,keepaspectratio]{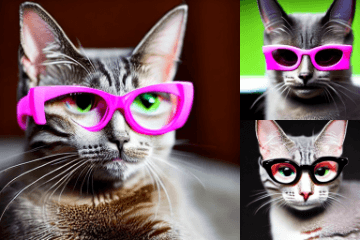}}
\newcommand{\catdbfiref}{\includegraphics[width=\colw,height=\colw,keepaspectratio]{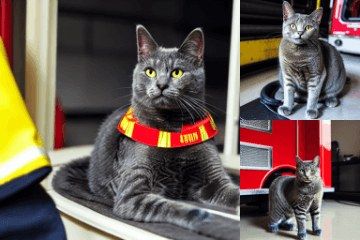}}

\newcommand{\dogddbmountain}{\includegraphics[width=\colw,height=\colw,keepaspectratio]{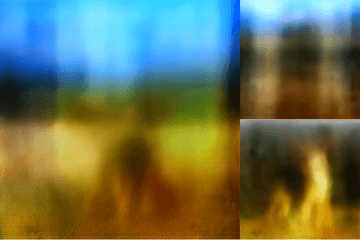}}
\newcommand{\dogddbblue}{\includegraphics[width=\colw,height=\colw,keepaspectratio]{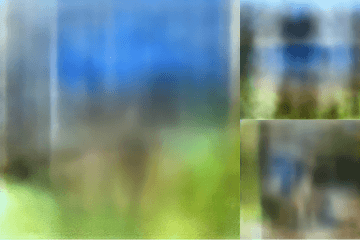}}
\newcommand{\dogddbsnow}{\includegraphics[width=\colw,height=\colw,keepaspectratio]{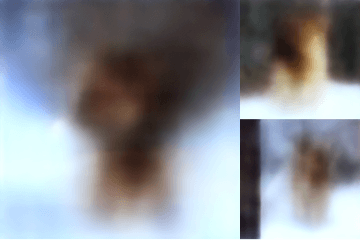}}

\newcommand{\catddbpink}{\includegraphics[width=\colw,height=\colw,keepaspectratio]{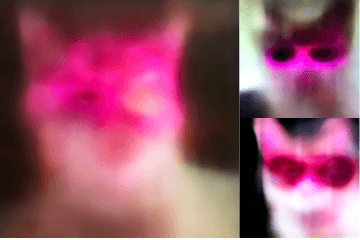}}
\newcommand{\catddbjungle}{\includegraphics[width=\colw,height=\colw,keepaspectratio]{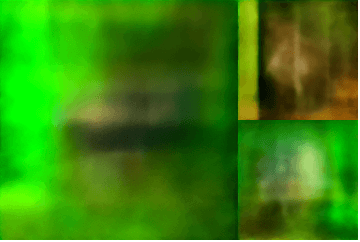}}
\newcommand{\catddbfiref}{\includegraphics[width=\colw,height=\colw,keepaspectratio]{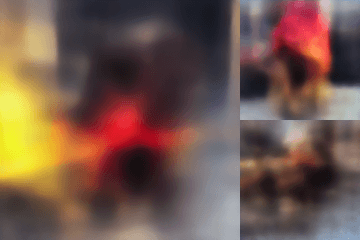}}

\newcommand{\vaseddbbeach}{\includegraphics[width=\colw,height=\colw,keepaspectratio]{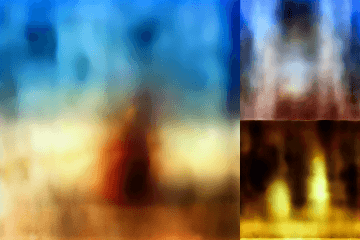}}
\newcommand{\vaseddbcobb}{\includegraphics[width=\colw,height=\colw,keepaspectratio]{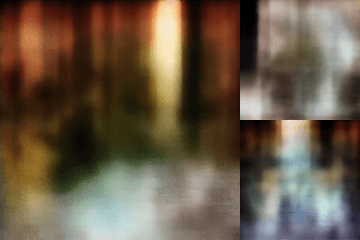}}
\newcommand{\vaseddbsunf}{\includegraphics[width=\colw,height=\colw,keepaspectratio]{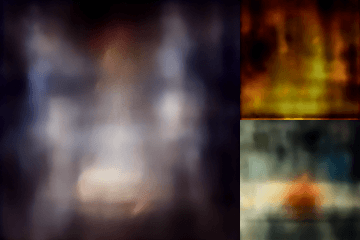}}

\newcommand{\vasedbdmdbeach}{\includegraphics[width=\colw,height=\colw,keepaspectratio]{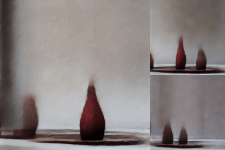}}
\newcommand{\vasedbdmdsun}{\includegraphics[width=\colw,height=\colw,keepaspectratio]{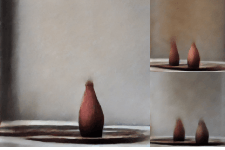}}
\newcommand{\vasedbdmdstreet}{\includegraphics[width=\colw,height=\colw,keepaspectratio]{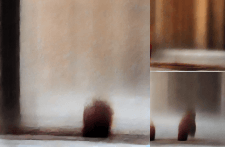}}

\newcommand{\catdpinkdbdmd}{\includegraphics[width=\colw,height=\colw,keepaspectratio]{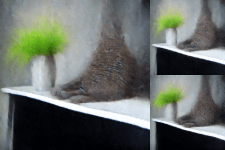}}
\newcommand{\catjungledbdmd}{\includegraphics[width=\colw,height=\colw,keepaspectratio]{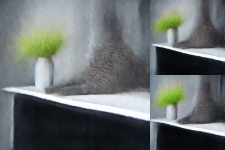}}
\newcommand{\catfirfdbdmd}{\includegraphics[width=\colw,height=\colw,keepaspectratio]{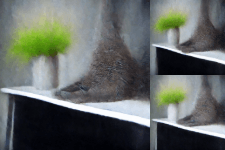}}

\newcommand{\dogmountaindbdmd}{\includegraphics[width=\colw,height=\colw,keepaspectratio]{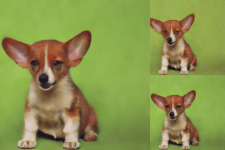}}
\newcommand{\dogsnowdbdmd}{\includegraphics[width=\colw,height=\colw,keepaspectratio]{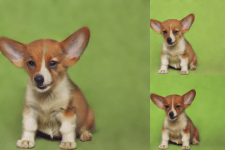}}
\newcommand{\dogbluedbdmd}{\includegraphics[width=\colw,height=\colw,keepaspectratio]{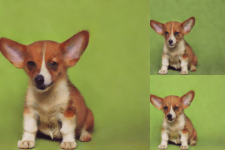}}

\newcommand{\dogpsomountain}{\includegraphics[width=\colw,height=\colw,keepaspectratio]{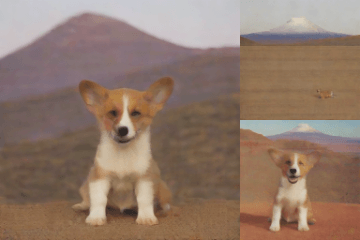}}
\newcommand{\dogpsoblue}{\includegraphics[width=\colw,height=\colw,keepaspectratio]{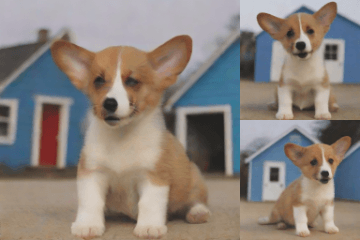}}
\newcommand{\dogpsosnow}{\includegraphics[width=\colw,height=\colw,keepaspectratio]{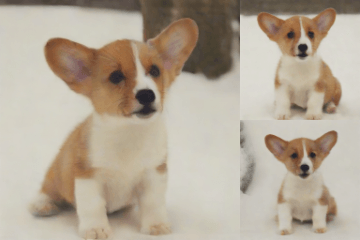}}

\newcommand{\catpsopink}{\includegraphics[width=\colw,height=\colw,keepaspectratio]{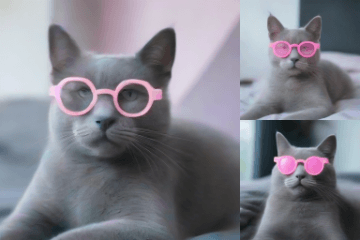}}
\newcommand{\catpsojungle}{\includegraphics[width=\colw,height=\colw,keepaspectratio]{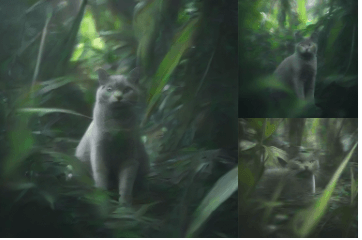}}
\newcommand{\catpsofiref}{\includegraphics[width=\colw,height=\colw,keepaspectratio]{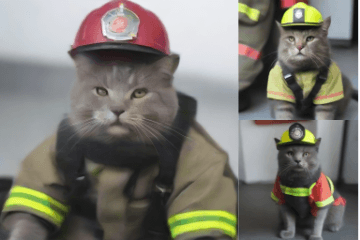}}

\newcommand{\vasepsobeach}{\includegraphics[width=\colw,height=\colw,keepaspectratio]{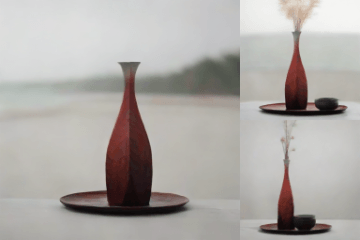}}
\newcommand{\vasepsocobb}{\includegraphics[width=\colw,height=\colw,keepaspectratio]{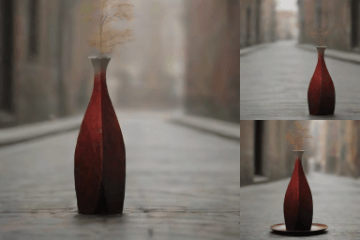}}
\newcommand{\vasepsosunf}{\includegraphics[width=\colw,height=\colw,keepaspectratio]{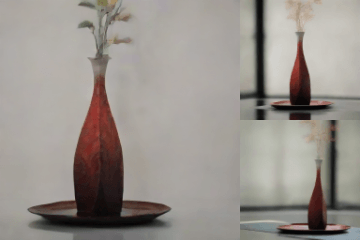}}

\newcommand{\backpackpso}{\includegraphics[width=\colw,height=\colw,keepaspectratio]{src/figures/picture_SDIG/backpack_pso.png}}

\newcommand{\wolfypso}{\includegraphics[width=\colw,height=\colw,keepaspectratio]{src/figures/picture_SDIG/wolfy_pso.png}}

\begingroup
\newcommand{\headsize}{\footnotesize}
\newcommand{\hdrplain}[1]{\makebox[\colw][c]{\headsize\bfseries #1}}

\newcommand{\threeprompt}[1]{\multicolumn{3}{l}{\scriptsize #1}}

\begingroup
\setlength{\arrayrulewidth}{0.8pt}%
\arrayrulecolor{black}%

\resizebox{\textwidth}{!}{
\begin{tabular}{
  >{\centering\arraybackslash}m{\colw} 
  >{\centering\arraybackslash}m{\colw} 
  *{5}{>{\centering\arraybackslash}m{\colw}}
}
\hdrplain{Target Set} &
\shortstack[c]{\headsize\bfseries \outlinebox{~\texttt{Uni-DAD}~}\\\headsize SDv1.5, NFE = 1} &
\shortstack[c]{\headsize\bfseries FT \\\headsize SDv1.5, NFE = 2$\times$50} &
\shortstack[c]{\headsize\bfseries \textcolor{red}{DMD2-FT}\\\headsize SDv1.5, NFE = 1} &
\shortstack[c]{\headsize\bfseries \textcolor{orange}{FT-DMD2}\\\headsize SDv1.5, NFE = 1} &
\shortstack[c]{\headsize\bfseries Turbo-PSO\\\headsize SDXL, NFE = 4} \\[2pt]
\midrule

\cattarget & \catunidadpinkg & \catdbpink & \catddbpink & \catdpinkdbdmd & \catpsopink  \\[+4pt]
{\scriptsize    } &\threeprompt{``a \texttt{prt} cat wearing pink glasses''} \\[+3pt]

{\scriptsize ``a \texttt{prt} cat''} & \catunidadjungle & \catdbjungle & \catddbjungle & \catjungledbdmd & \catpsojungle  \\[+4pt]
{\scriptsize } &\threeprompt{``a \texttt{prt} cat in the jungle''} &\\[-10pt]

\end{tabular}
} 
\arrayrulecolor{black}%
\endgroup

\caption{Qualitative comparison for SDP, adapting SDV1.5~\cite{rombach2022high} to the DreamBooth~\cite{ruiz2023dreambooth} \emph{cat2} subject, evaluated on accessorization and re-contextualization prompts. See additional results on other subjects (\emph{dog6, vase}) and prompts in Fig.~\ref{fig:A_SDP_qualitative}. Zoom in for details.}
\label{fig:4_SDP_qualitative}
\endgroup

\end{figure*}

%% file: src/tables/4_SDIG_Quantitative_Overall.tex
\begin{table*}[!t]
\centering
\footnotesize

\setlength{\tabcolsep}{0pt}
\renewcommand{\arraystretch}{1.08}

\caption{Comparison of quality (DINO$\uparrow$, CLIP-I$\uparrow$, CLIP-T$\uparrow$) and diversity (Intra-LPIPS$\uparrow$, Inter-LPIPS$\uparrow$) for SDP across methods, evaluated on the DreamBooth benchmark (30 subjects, 25 prompts)~\cite{ruiz2023dreambooth}. \textbf{Best} and \underline{second best} distilled method at NFE=1.}
\label{tab:sdp_overall}

\resizebox{\textwidth}{!}{%
\begin{tabular}{@{}
L{2.0cm}
L{3.0cm}
C{1.6cm}
C{1.6cm}
C{1.7cm}
C{1.7cm}
C{1.7cm}
C{2.75cm}
C{2.75cm}
@{}}
\toprule

& & & & \multicolumn{3}{c}{\textbf{Quality}} & \multicolumn{2}{c}{\textbf{Diversity}} \\
\cmidrule(lr){5-7}\cmidrule(lr){8-9}

\textbf{Variant} & \textbf{Method} & \textbf{NFE}$\downarrow$ & \textbf{1-stage} & \textbf{DINO}$\uparrow$ & \textbf{CLIP-I}$\uparrow$ & \textbf{CLIP-T}$\uparrow$ & \textbf{Intra-LPIPS}$\uparrow$ & \textbf{Inter-LPIPS}$\uparrow$ \\
\midrule

\emph{Non-distilled} & FT~\cite{ruiz2023dreambooth} & $2\times50$ & $\checkmark$ & 0.58 & 0.77 & 0.32 & 0.67$\pm$0.08 & 0.73$\pm$0.06 \\
\midrule

& Turbo-PSO$_{\text{SDXL}}$~\cite{miao2024tuning} & 4 &  & 0.50 & 0.70 & 0.30 & 0.42$\pm$0.07 & 0.60$\pm$0.08 \\[-1pt]

\cmidrule{2-9}

& DMD2-PSO$_{\text{SDv1.5}}$~\cite{miao2024tuning} & \textbf{1} &  & 0.14 & 0.56 & 0.23 & 0.07$\pm$0.02 & 0.11$\pm$0.02 \\[-1pt]

\emph{Distilled} & \textcolor{red}{DMD2-FT} & \textbf{1} &  & 0.20 & 0.61 & \underline{0.26} &
\multicolumn{1}{>{\columncolor{red!10}\centering\arraybackslash}m{2.75cm}}{\textbf{0.58}$\pm$0.07} &
\multicolumn{1}{>{\columncolor{red!10}\centering\arraybackslash}m{2.75cm}}{\textbf{0.70}$\pm$0.09} \\

& \textcolor{orange}{FT-DMD2} & \textbf{1} &  &
\multicolumn{1}{>{\columncolor{orange!10}\centering\arraybackslash}m{1.7cm}}{\textbf{0.57}} &
\multicolumn{1}{>{\columncolor{orange!10}\centering\arraybackslash}m{1.7cm}}{\textbf{0.75}} &
0.25 & 0.22$\pm$0.04 & 0.25$\pm$0.07 \\

\rowcolor{black!6}
& \texttt{Uni-DAD} & \textbf{1} & $\checkmark$ & \underline{0.47} & \underline{0.73} & \textbf{0.29} & \underline{0.51}$\pm$0.09 & \underline{0.59}$\pm$0.09 \\
\bottomrule

\end{tabular}%
}
\vspace{-8pt}
\end{table*}

%% file: sec/5_conclusion.tex
\section{Conclusion}
\label{sec:conclusion}

We introduced \texttt{Uni-DAD}, a single-stage pipeline that unifies diffusion model distillation and adaptation for fast few-shot image generation. By combining a dual-domain DMD objective with a multi-head GAN loss, it preserves transferable source-domain knowledge while improving target-domain realism under scarce data. Evaluated across two benchmarks, FSIG and SDP, and using different diffusion backbones, \texttt{Uni-DAD} delivers better or comparable quality to SoTA adaptation methods even with $\leq 4$ sampling steps, often surpassing two-stage pipelines in quality and diversity. Overall, our results suggest that distillation and adaptation of DMs need not be treated as separate stages, opening a path toward fast and high-quality image generation under scarce data and non-trivial domain shifts.

\noindent\textbf{Limitations and Future Work.}
\texttt{Uni-DAD} inherits the sensitivity of GAN training, including hyperparameter tuning and the overfitting risk in small target sets. Although it significantly reduces sampling cost, training remains more expensive than standard adaptation alone. Future work will explore more parameter-efficient variants, improved scheduling and learning of the dual-domain weighting, and extensions to larger backbones and other modalities, including video and audio diffusion models.

\section{Acknowledgements}
This research was supported by the Natural Sciences and Engineering Research Council of Canada, and the Digital Research Alliance of Canada.




%% file: sec/X_suppl.tex
\clearpage
\setcounter{page}{1}
\maketitlesupplementary

\begin{center}
  \begin{minipage}{1\linewidth}
    \small
    \hrule\vspace{0.6em}
    \begin{center}
      \textbf{Table of Contents}
    \end{center}
    \hrule
    \vspace{0.7em}

    \SuppTOCLine{\textbf{~\ref{sec:A_methodology}\ ~~ Proposed Method Contd.}}{\pageref{sec:A_methodology}}
    \SuppTOCLine{\quad \ref{sec:A_algorithm}\ ~~ \texttt{Uni-DAD} Training Iteration}{\pageref{sec:A_algorithm}}
    \SuppTOCLine{\quad \ref{sec:A_SDP_methodology}\ ~~ Adapting \texttt{Uni-DAD} to SDP}{\pageref{sec:A_SDP_methodology}}

    \vspace{0.4em}

    \SuppTOCLine{\textbf{~\ref{sec:A_training_evaluation}\ ~~ Results and Discussion Contd.}}{\pageref{sec:A_training_evaluation}}
    \SuppTOCLine{\quad \ref{sec:A_details_FSIG}\ ~~ FSIG Training Details}{\pageref{sec:A_details_FSIG}}
    \SuppTOCLine{\quad \ref{sec:A_FSIG_additional}\ ~~ FSIG Additional Generated Samples}{\pageref{sec:A_FSIG_additional}}
    \SuppTOCLine{\quad \ref{sec:A_ablations_FSIG}\ ~~ FSIG Ablations Contd.}{\pageref{sec:A_ablations_FSIG}}

    \vspace{0.25em}

    \SuppTOCLine{\quad \ref{sec:A_details_SDP}\ ~~ SDP Training Details}{\pageref{sec:A_details_SDP}}
    \SuppTOCLine{\quad \ref{sec:A_SDP_additional}\ ~~ SDP Additional Generated Samples}{\pageref{sec:A_SDP_additional}}
    \SuppTOCLine{\quad \ref{sec:A_ablations_SDP}\ ~~ SDP Ablations Contd.}{\pageref{sec:A_ablations_SDP}}

    \vspace{0.25em}

    \SuppTOCLine{\quad \ref{sec:A_style_transfer}\ ~~ \texttt{Uni-DAD} in Style Transfer}{\pageref{sec:A_style_transfer}}

    \vspace{0.3em}\hrule
  \end{minipage}
\end{center}

\section{Proposed Method Contd.}
\label{sec:A_methodology}

\noindent\textbf{Why a source score is useful.} The source teacher $\epsilon^{\text{src}}$, trained on large and diverse data, can be viewed as a general image-manifold approximator~\cite{bahram2025dogfit}. Moreover, it has been shown that diffusion models operating in noisy space can generalize across distributions and denoise out-of-domain inputs~\cite{deja2022analyzing,meng2021sdedit}. Consequently, $\epsilon^{\text{src}}$ offers a stable score around $x_t$ that regularizes $G$ and helps preserve general information shared between the source and target domains.

\subsection{\texttt{Uni-DAD} Training Iteration}
\label{sec:A_algorithm}
Alg.~\ref{alg:uni_dad} describes the training of \texttt{Uni-DAD} at each iteration.




\input{src/algorithms/algorithm}

\subsection{Adapting \texttt{Uni-DAD} to SDP}
\label{sec:A_SDP_methodology}

\texttt{Uni-DAD} provides an efficient and high-quality pipeline for SDP using text-conditioned DMs~\cite{rombach2022high, podell2023sdxl}. In this setting, the goal is to learn a new subject identity from only a handful of images and reproduce it faithfully across diverse textual prompts. Here, we outline additional methodology details for adapting \texttt{Uni-DAD} to this task.

\noindent\textbf{Conditioning on Subject Prompts.} SDP requires conditioning the DM on textual prompts that specify the subject identity. \texttt{Uni-DAD} naturally extends to this setting by incorporating prompt conditions into all score evaluations. Let $c$ denote the \emph{subject prompt}, formatted as “a [rare token] [class noun]”, where the rare token uniquely identifies the target subject and the class noun specifies the broader semantic class (e.g., “dog”, “cat”, "vase"). This prompt is provided to the models that undergo learning, i.e., the student generator $G$, the fake teacher $\epsilon^{\text{fk}}$, and the target teacher $\epsilon^{\text{trg}}$. Injecting $c$ ensures that these models associate the rare token with the subject identity being learned.

\noindent\textbf{Class-Prior Prompts.} To maintain generality and prevent overfitting, we additionally define a \emph{class-prior prompt} $c^{\text{prior}} = \text{``a [class noun]''}$, which is fed to the frozen source teacher $\epsilon^{\text{src}}$. Since $\epsilon^{\text{src}}$ has not been trained on the specific subject, $c^{\text{prior}}$ enables it to produce class-consistent but subject-agnostic guidance. This separation of prompts is crucial: $\epsilon^{\text{src}}$ continues to act as a diversity-inducing regularizer, stabilizing identity learning by preserving shared class-level structure.

\noindent\textbf{SDP Components.} All components of \texttt{Uni-DAD} (i.e., dual-domain DMD, fake and target teachers, and the multi-head GAN) operate unchanged for SDP except for prompt conditioning at each time-step. The dual-domain DMD now aligns conditional distributions, guiding the student to preserve general class semantics via $\epsilon^{\text{src}}(\cdot|c^{\text{prior}})$ while adapting to the personalized subject via $\epsilon^{\text{trg}}(\cdot|c)$. Likewise, the GAN discriminator receives $c$ to encourage realistic subject-specific details at multiple feature scales.

\section{Results and Discussion Contd.}
\label{sec:A_training_evaluation}

\subsection{FSIG Training Details}
\label{sec:A_details_FSIG}

\begin{figure}[t]
  \centering
  \includegraphics[width=\linewidth]{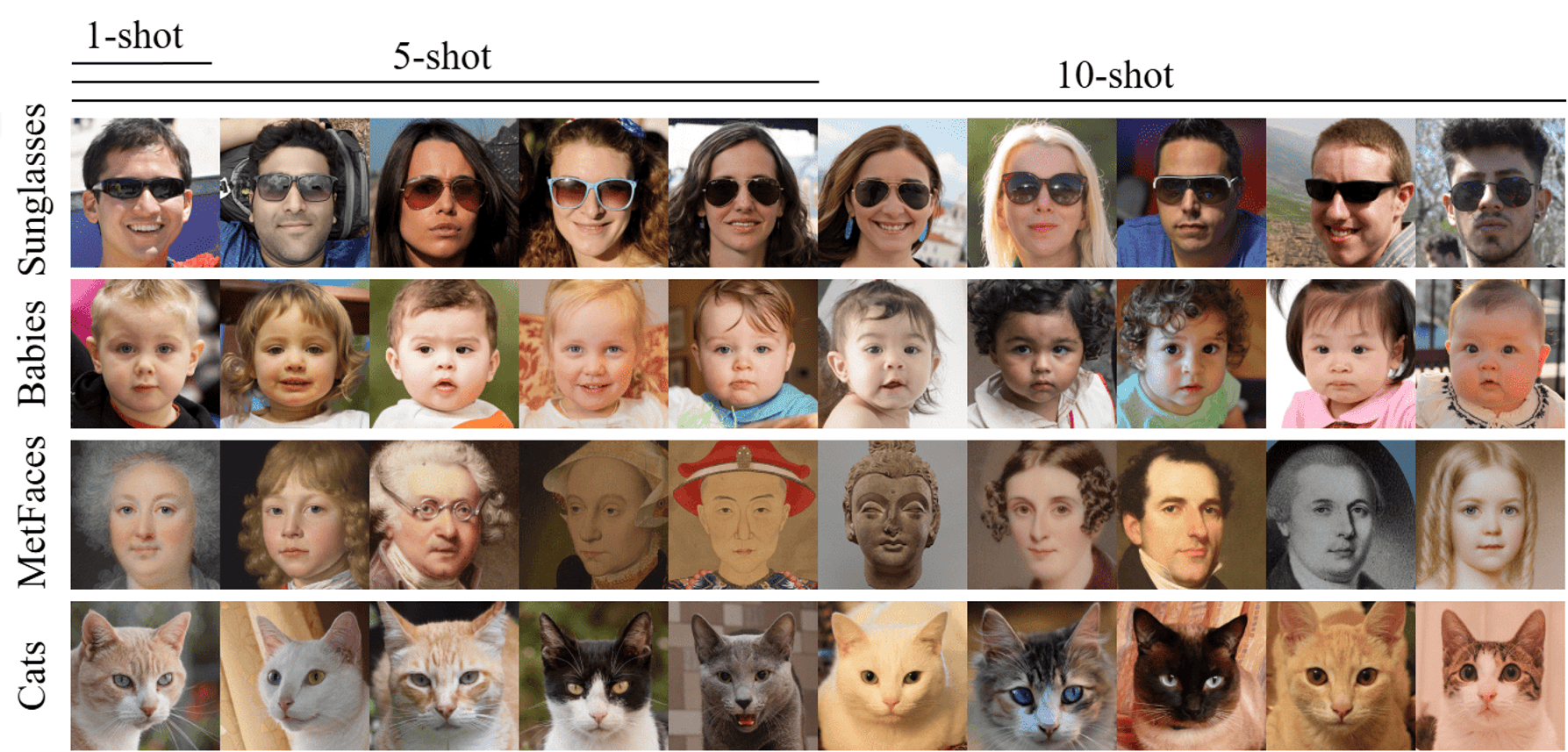}
  \caption{Representative target sets used in our experiments. Rows: domains. Column groups: \(k\)-shot setting (\(k\in\{1,5,10\}\)).}
  \label{fig:A_FSIG_targets}
\end{figure}

\noindent\textbf{Selected images for the $k$-shot ablation.}
Figure~\ref{fig:A_FSIG_targets} shows representative training images for ablating the number of shots \(k\in\{1,5,10\}\) per target domain in Table~\ref{tab:4_kshot_fid}. 

\vspace{4px}

\noindent\textbf{Global Details.} We include details of \texttt{Uni-DAD}, DMD2-FT, and CRDI~\cite{cao2024few}. All models are trained with image size $256 \times 256$ and each dataset is resized from $1024 \times 1024$ to this resolution for training. Since the Inception network expects $299 \times 299$ input images for FID calculation, following common practice, we adopt the $1024 \to 256 \to 299$ resizing
pipeline for consistency of evaluation with prior work~\cite{cao2024few}. We report the best FID\footnote{FID: \url{https://github.com/bioinf-jku/TTUR}} observed during training over 20K iterations and its corresponding Intra-LPIPS\footnote{Intra-LPIPS: \url{https://github.com/YuCao16/CRDI}}.

\vspace{4px}

\noindent\textbf{\texttt{Uni-DAD} Training.} Training is conducted over one 80GB H100 GPU for 2 to 3 hours~\ref{tab:A_Compute_cost}. When using $e^{\text{trg}}$, We set $a=0.25$ for close domains (Babies, Sunglasses), and $a=0.75$ for distant ones (MetFaces, Cats). Furthermore, we set $\lambda_{\text{GAN}}^{G}=0.01$ and $\lambda_{\text{GAN}}^{D}=0.03$ for all experiments. All experiments are performed with $\text{NFE}=3$ unless specified otherwise. The generator update ratio is set to 5. For the multi-head GAN discriminator, for each feature map of dimension $C \times H \times W$, we apply a single $1\times1$ convolution to project the $C$ channels to a single-channel map, followed by a global average pooling to obtain a scalar logit. This directly aggregates information from multiple resolutions. For the single-head GAN classifier, we use a deep bottleneck branch based on DMD2~\citep{yin2024improved}. We use a batch size of one, mixed-precision (bf16), and random horizontal flipping augmentation. The learning rate is $2e^{-6}$ for all models.

\vspace{4px}

\noindent\textbf{DMD2-FT Training.} 
The DMD2 distilled model generates FFHQ-aligned images and attains FID@5k of 24.80. For fine-tuning the distilled model, we test two possible design choices of training-time time-step sampling: (i) Only sampling time-steps based on $\text{NFE}$ (e.g., $\text{NFE}=3,~t\in \{333,666,1000\}$, and (ii) Sampling from the whole possible steps $T\in\{1,...,1000\}$. We observe similar behavior in both cases and choose the first option. We use a batch size of one, mixed-precision (bf16), and random horizontal flipping augmentation. The learning rate is $2e^{-6}$ for both stages. FT-DMD2 follows the same hyperparameter configuration as DMD2-FT.

\vspace{4px}

\noindent\textbf{CRDI Training.}
We train CRDI~\citep{cao2024few} using the authors' released code and configurations\footnote{CRDI: \url{https://github.com/YuCao16/CRDI}}. We use four A100 GPUs to match their batch size of 10 and train for roughly 1 GPU hour. On closer domains (Sunglasses and Babies), following the authors' guidance, we set $t_\text{start}=5$, $t_\text{end}=20$, and $\text{num\_gradient}=15$. On MetFaces, we use $t_\text{start}=5$, $t_\text{end}=15$, and $\text{num\_gradient}=10$ as suggested. However, the FID that we attain in this case is different what they report. We include both our results and theirs. On Cats, we adopt the same configuration as MetFaces.

\subsection{FSIG Additional Generated Samples}
\label{sec:A_FSIG_additional}
\begin{figure*}[t]
    \centering
    \includegraphics[width=\textwidth]{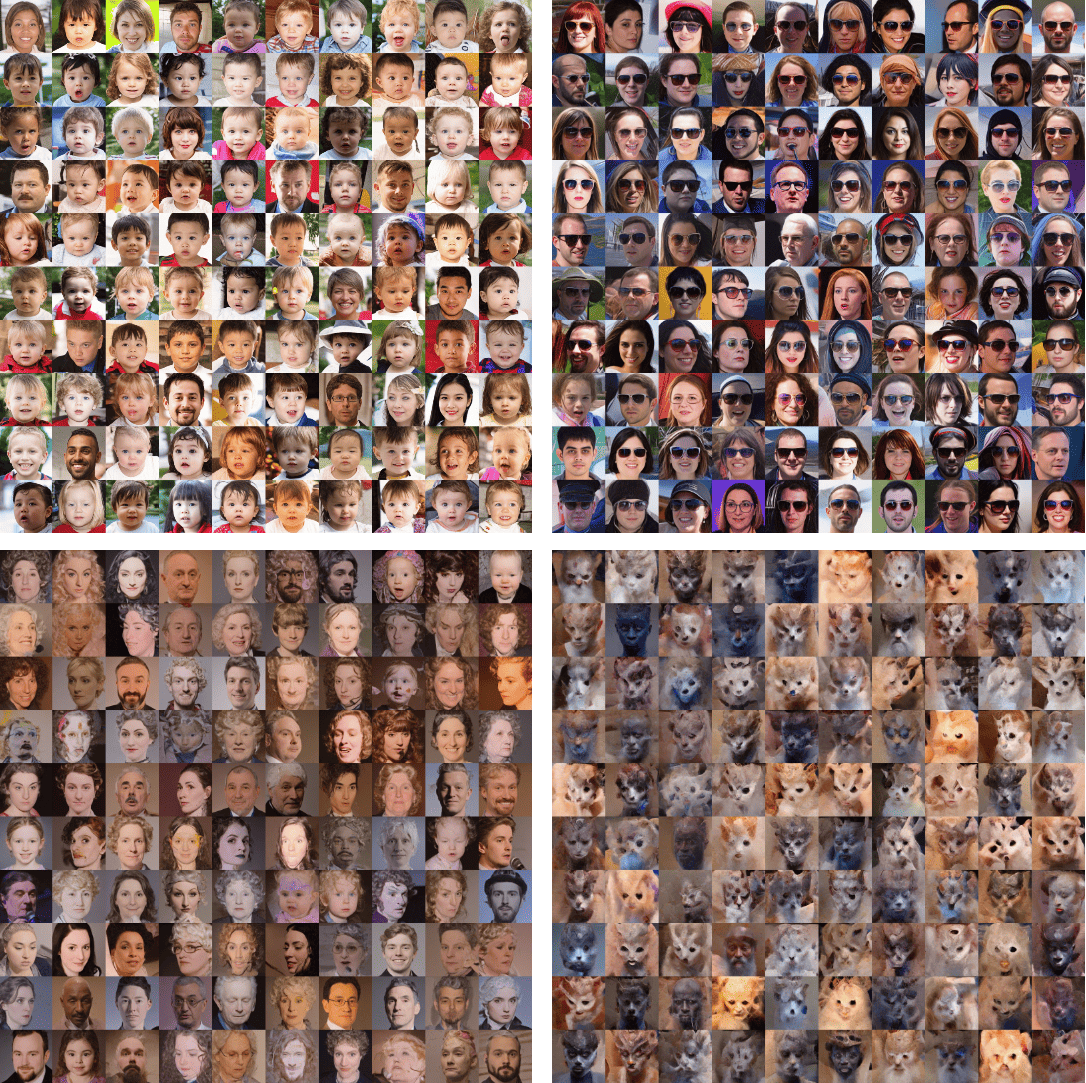}
    \caption{Additional generated samples of FSIG using \texttt{Uni-DAD} \textbf{without} $\epsilon^{\text{trg}}$ ($a=0$ for all domains).}
    \label{fig:A_all_qualitative_source}
\end{figure*}

\begin{figure*}[t]
    \centering
    \includegraphics[width=\textwidth]{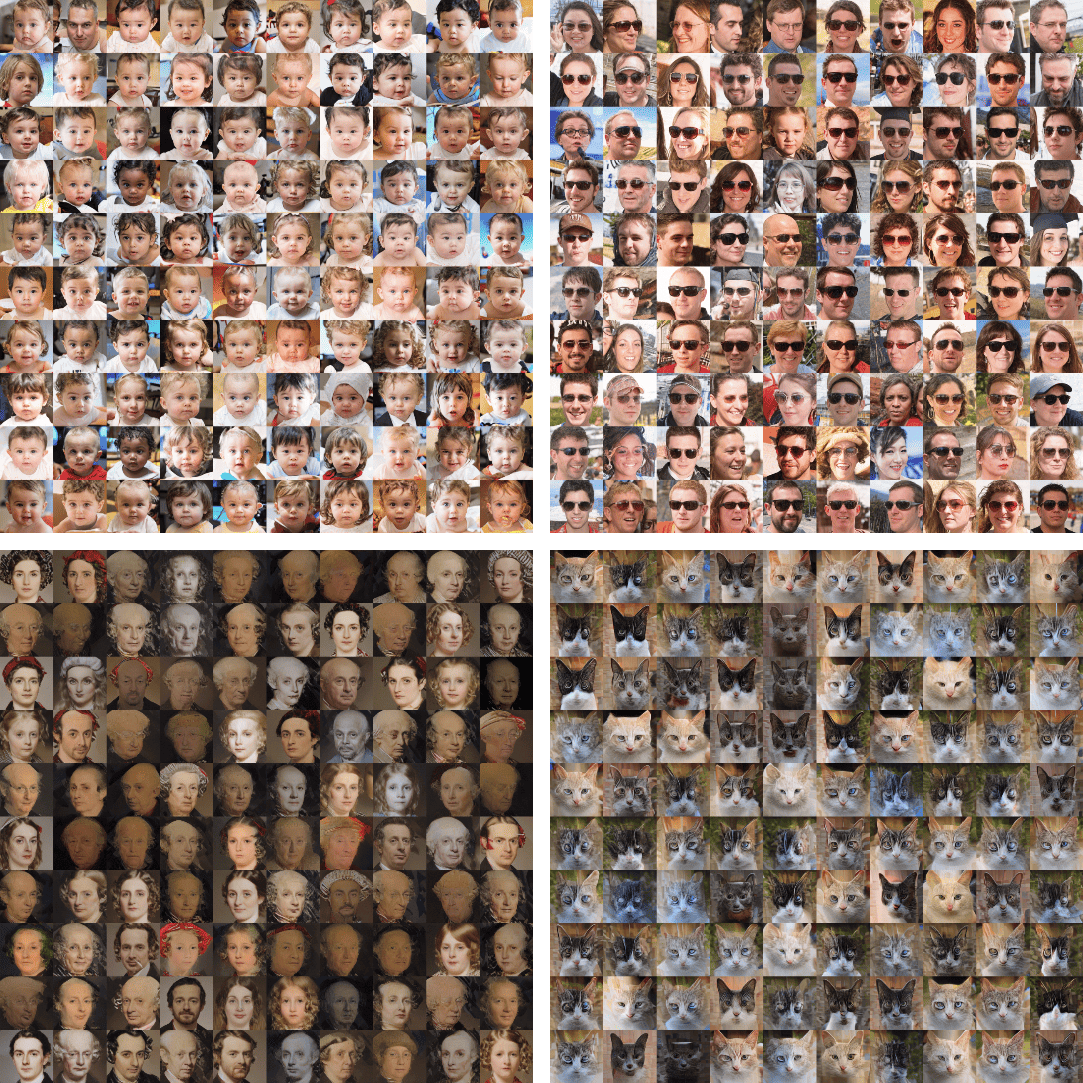}
    \caption{Additional generated samples of FSIG using \texttt{Uni-DAD} \text{with} $\epsilon^{\text{trg}}$ ($a=0.25$ for Babies/Sunglasses and $a=0.75$ for MetFaces/Cats).}
    \label{fig:A_all_qualitative_source_target}
\end{figure*}

Figs.~\ref{fig:A_all_qualitative_source} and~\ref{fig:A_all_qualitative_source_target} show 100 additional samples generated by \texttt{Uni-DAD} on Babies, Sunglasses, MetFaces, and Cats, without and with target teacher $\epsilon^{\text{trg}}$, respectively. Without $\epsilon^{\text{trg}}$, the model fully adapts to close domains (Babies and Sunglasses), and adapts to the style of the distant target domains (MetFaces and Cats). The inclusion of the target teacher allows higher fidelity to the structure of the distant domains, at the cost of slight diversity reduction.

\subsection{FSIG Ablations Contd.}
\label{sec:A_ablations_FSIG}

\input{src/tables/4_FSIG_kshot}

\input{src/tables/4_FSIG_Ablation_Components}
\input{src/tables/4_FSIG_Ablation_Init}

\input{src/tables/4_FSIG_Ablation_GANloss}

\noindent \textbf{(b) Target Set Size and $\text{NFE}$.}
Tab.~\ref{tab:4_kshot_fid} ablates quantitative performance over $\text{NFE}\in \{1,2,3,4\}$ in 1/5/10-shot settings. Increasing $\text{NFE}$ improves FID up until $\text{NFE}=3$, with little to no gain at $\text{NFE}=4$. This highlights the effectiveness of our method as a few-step few-shot image generator. With $\text{NFE}=3$, \texttt{Uni-DAD} consistently attains lower FID than CRDI in all settings, indicating its robustness across different few-shot regimes.

\vspace{4px}

\noindent \textbf{(c) Component Analysis.} 
Tab.~\ref{tab:4_components} isolates the impact of the dual-domain DMD and the multi-head GAN on FID. While a single-head GAN outperforms our multi-head design when DMD is absent, the multi-head GAN becomes increasingly beneficial once paired with the DMD losses. The implication is that enforcing realism at multiple feature levels helps mitigate overfitting in few-shot contexts. Moreover, using DMD without the GAN often causes training instability and drift, which is mitigated by the target realism signals provided by the GAN. Overall, all components of our approach jointly improve generation quality.

\vspace{4px}

\noindent \textbf{(d) Available checkpoints.} 
Tab.~\ref{tab:4_init_ablation} ablates the effect of having different pre-trained checkpoints at the start of \texttt{Uni-DAD} training. In practice, one may have access to a distilled source model (\emph{Pre-distilled $G$}), an adapted DM in the target domain (\emph{Pre-adapted $\epsilon^{\text{trg}}$}), or both. \texttt{Uni-DAD} is checkpoint-agnostic: an adapted DM's weights can replace $\epsilon^{\text{trg}}$ with no online training needed, and a distilled source model can initialize the student. This makes \texttt{Uni-DAD} applicable both as an adaptation method for distilled models and as a distillation method for adapted models.

\vspace{4px}

\noindent \textbf{(e) Type of GAN Loss.} 
Tab.~\ref{tab:4_gan_loss_ablation} ablates the GAN loss. Among hinge~\cite{lim2017geometric}, least-squares (LSGAN)~\cite{mao2017least}, Wasserstein (WGAN)~\cite{arjovsky2017wasserstein}, and binary cross-entropy (BCE)~\cite{goodfellow2014generative}, BCE yields the best FID with the multi-head GAN on both Babies and MetFaces. WGAN is noticeably less stable in our few-shot setting, showing sudden loss spikes and occasional divergence, likely due to its sensitivity to hyperparameters and discriminator overfitting. By contrast, BCE, hinge, and LSGAN train more consistently, with BCE giving the strongest overall results. The results further indicate that under stable training, the multi-head discriminator often surpasses the single-head variant, indicating that discrimination at multiple feature levels improves robustness in few-shot adaptation.

\subsection{SDP Training Details}
\label{sec:A_details_SDP}

\noindent\textbf{\texttt{Uni-DAD SDP} Training.} 
All models are trained with a learning rate of $5\times10^{-6}$. The multi-head GAN uses discriminator and generator weights of $\lambda_{\text{GAN}}^{D}=0.01$ and $\lambda_{\text{GAN}}^{G}=0.001$, respectively. The update ratio for $G$ and $\epsilon^{\text{trg}}$ is set to 10. The student generator $G$ is initialized from the DMD2 pre-distilled SDv1.5 weights~\footnote{DMD2 weights: \url{https://github.com/tianweiy/DMD2}}. We set the DMD weighting factor to $a=0.75$, matching the distant-domain configuration used for FSIG. \texttt{Uni-DAD} is trained for 5k iterations on a single H100 GPU ($\approx$50 GB memory usage), with best generations typically appearing between 4k–5k steps. Training time is $\approx$ 1.6 hour per subject.

\vspace{4px}

\noindent\textbf{FT Training.}
For the \textit{Non-distilled} FT baseline, we apply DreamBooth-style fine-tuning~\cite{ruiz2023dreambooth}. We follow prior preservation training by generating 1,000 samples of the form “a [class noun]” using SDv1.5. We fine-tune for 800 iterations per instance, using a batch size of 1 and a fixed learning rate of $5\times10^{-6}$. DreamBooth commonly uses 400–1,200 steps depending on the subject~\cite{kumari2023custom, miao2024tuning, Ram_2025_WACV, wei2023elite}. we find 800 steps of training to be sufficient in reproducing the authors' reported quality across subjects.

\vspace{4px}

\noindent\textbf{DMD2-FT Training.} 
To construct this two-stage \textit{distilled} baseline, we initialize the FT pipeline using DMD2 SDv1.5 checkpoints and fine-tune the student with DreamBooth. We perform one-step sampling following DMD2-distilled generation.

\vspace{4px}

\noindent\textbf{FT-DMD2 Training.} To construct this two-stage \textit{distilled} baseline, we initialize the DMD2 pipeline using DreamBooth SDv1.5 fine-tuned checkpoints. We perform one-step sampling following DMD2-distilled generation.

\vspace{4px}

\noindent\textbf{PSO Training.}
PSO operates on the SDXL-Turbo \textit{distilled} backbone~\cite{sauer2024adversarial} and fine-tunes it using pairwise sample optimization~\cite{miao2024tuning}. It performs 4-step sampling following SDXL-Turbo-distilled generation. We use the official training setup for each subject. All models are evaluated after 800 training steps\footnote{PSO: \url{https://github.com/ZichenMiao/Pairwise_Sample_Optimization}}.

\subsection{SDP Additional Generated Samples}
\label{sec:A_SDP_additional}

\begin{figure}[!t]
  \centering
  \includegraphics[width=\linewidth]{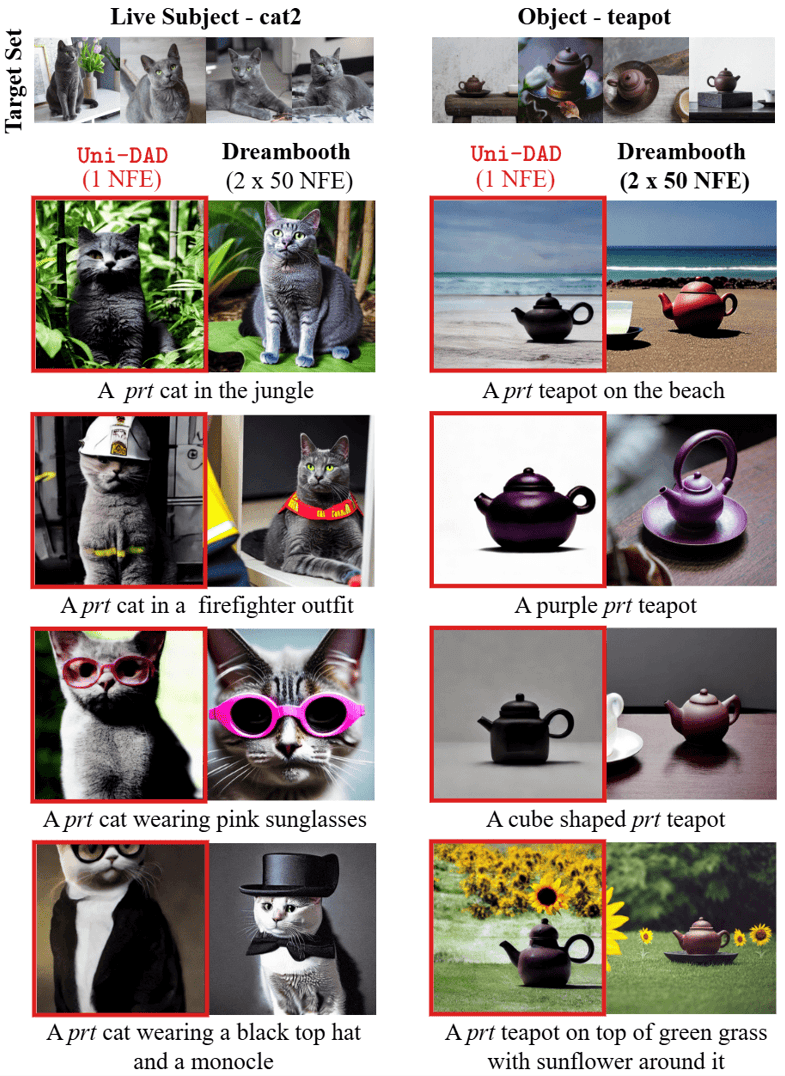}  
  \caption{
  \outlinebox{~\texttt{Uni-DAD}~} vs. DreamBooth~\cite{ruiz2023dreambooth} for SDP. \emph{prt} is used as a rare token to support learning of novel target subjects (\emph{cat2} and \emph{teapot}). $\text{NFE}$ is reduced from 100 to 1. 
}
  \label{fig:overviewSDP}
\end{figure}

Fig.~\ref{fig:overviewSDP} provides a qualitative comparison between \emph{Distilled} \texttt{Uni-DAD} (NFE$~=1$) and \emph{Non-distilled} DreamBooth ~\cite{ruiz2023dreambooth} (NFE$~=2\times50$) for \emph{cat2} and \emph{teapot} using diverse prompts that re-contextualize the personalized instance, add accessories, or modify its properties. \texttt{Uni-DAD} preserves instance identity and follows prompts closely. Our model, exhibits slightly lower diversity that is commonly observed in distilled models and understandable given its heavy distillation~\citep{gandikota2025distilling}. 
Fig.~\ref{fig:A_SDP_qualitative} presents a comprehensive qualitative comparison between \texttt{Uni-DAD} and other SoTA SDP methods under diverse DreamBooth~\cite{ruiz2023dreambooth} prompts and subjects.

\subsection{SDP Ablations Contd.}
\label{sec:A_ablations_SDP}

\begin{figure}[!t] 
  \centering
  \includegraphics[width=\linewidth]{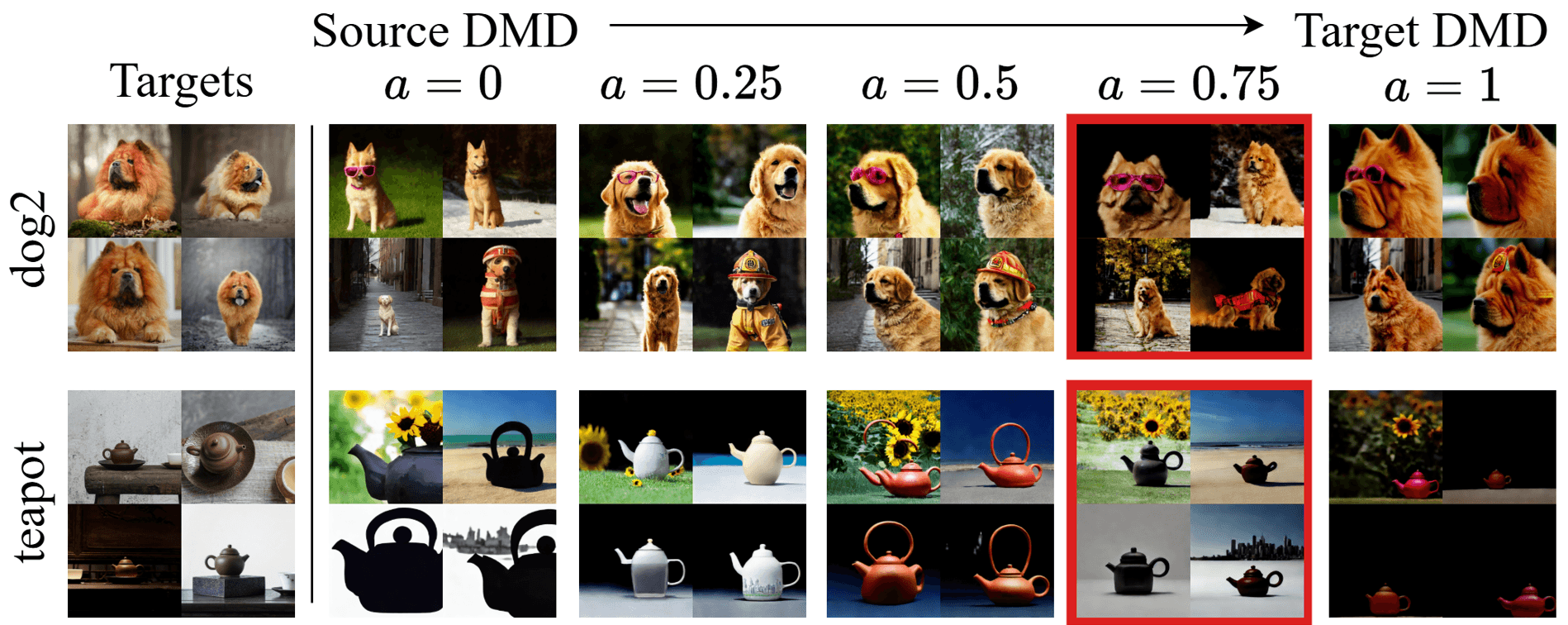}
  \vspace{-18pt}
  \caption{Qualitative ablation of the dual-domain DMD weighting factor $a$ for SDP across prompts on a live subject and an object.}
\label{fig:SDP_alpha}
\end{figure}

\noindent\textbf{(g) Quantitative Diversity Analysis.}
Fig.~\ref{fig:diversity_qualitative} qualitatively illustrates two diversity criteria for the \emph{dog7} subject. The first row reflects \emph{intra-prompt} diversity: for a fixed prompt, the model should generate varied but coherent outputs rather than repeat a single memorized configuration. The second row reflects \emph{inter-prompt} diversity: across prompts, it should respond to semantic changes while preserving the target identity. FT-DMD2 shows limited diversity in both settings, often reverting to nearly identical layouts, whereas DMD2-FT becomes too blurred for meaningful diversity. Turbo-PSO responds to prompt changes, but with more limited variation in pose and composition. In contrast, \texttt{Uni-DAD SDP} varies composition and appearance within a prompt and adapts clearly across prompts, while remaining visually consistent. These observations align with the Intra-LPIPS and Inter-LPIPS trends in Tab.~\ref{tab:sdp_overall}, supporting the stronger diversity-quality balance of \texttt{Uni-DAD} among \emph{Distilled} methods.

\input{src/tables/4_SDIG_qualitative_diversity}

\begin{figure*}[!t]
  \centering
  \includegraphics[width=\textwidth]{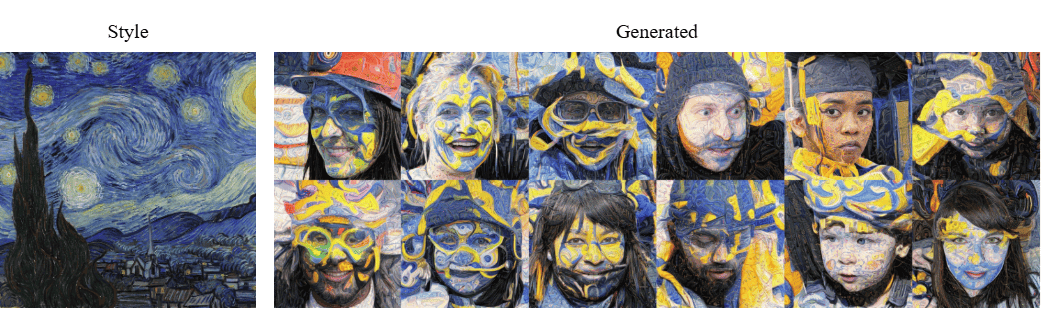}
  \caption{Style transferred images generated by \texttt{Uni-DAD}.}
  \label{fig:A_Vangogh}
\end{figure*}

\vspace{35px}

\noindent\textbf{(f) $a$ Coefficient.}
In Fig.~\ref{fig:SDP_alpha}, we qualitatively evaluate the value of $a$ on a live subject (dog2) and an object (teapot). On full the DreamBooth benchmark~\cite{ruiz2023dreambooth}, over all live subjects, $a$ of $\{0,0.25,0.5,0.75,1\}$ yields average CLIP-I$\uparrow$ of 0.771~/~0.794~/~\textbf{0.798}~/~0.789~/~0.785 and over all objects, it yields 0.652~/~0.656~/~0.661~/~\textbf{0.674}~/~0.657. Despite different optima ($a=0.5$ vs. $a=0.75$), we set $a=0.75$ for all SDP experiments based on visual inspection without target-specific tuning.

\FloatBarrier

\subsection{\texttt{Uni-DAD} in Style Transfer}
\label{sec:A_style_transfer}

\input{src/tables/4_SDIG_qualitative}

Our method can also be utilized as a one-shot style transfer technique, without requiring a target teacher ($\epsilon^{\text{trg}}$). As an example, we transfer and distill the FFHQ source model using the style of \textit{``The Starry Night'' by Vincent van Gogh}. As shown in Fig.~\ref{fig:A_Vangogh}, \texttt{Uni-DAD} successfully transfers the artistic style while preserving underlying facial structures and diversity in generation. This suggests that, as long as the target domain does not introduce major structural changes and differs with the source domain primarily in style, \texttt{Uni-DAD} can adapt to the new style using only the GAN branch as a driving force, while $\text{DMD}^{\text{src}}$ maintains consistency with the original source distribution. Babies and Sunglasses are two such cases, as they correspond to specialized subsets of the broader FFHQ distribution.

\stopsuppcontent

%% file: src/algorithms/algorithm.tex
\begin{algorithm}[t]
\caption{\texttt{Uni-DAD} Training Iteration}
\label{alg:uni_dad}
\DontPrintSemicolon
\SetKwInOut{Input}{Input}
\SetKwInOut{Output}{Output}

\Input{Source teacher $\epsilon^{\text{src}}$, Optional target teacher $\epsilon^{\text{trg}}$, Target set $Y=\{y\}$, Weight factor $a$, Training $step$, Update $ratio$}
\vspace{2pt}
\Output{Student $G$ (Adapted and Distilled)}
\vspace{3pt}
\setassignlhs{$\mathcal{L}_{\text{fk},D}$}

$\epsilon^{\text{fk}}\leftarrow\epsilon^{\text{src}}$ \\
\If{$\epsilon^{\text{trg}}==\varnothing$}{
$\epsilon^{\text{trg}}\leftarrow\epsilon^{\text{src}};~train\_target\leftarrow\textbf{True}$
}


\vspace{3pt}
\tcp{Prepare data}\vspace{2.5pt}
$t \sim \mathcal{U}\{0.02T, 0.98T\}$ \\
$(z,\epsilon) \sim \mathcal{N}(0,I)$ \\
\assignlinea{$y_t$}{$q~(y_t \mid y),~ y\sim Y$ \hfill \tcp{real}}
\assignlinea[]{$x_t$}{$q~(x_t \mid x),~ x\leftarrow G~(z)$ \hfill \tcp{fake}}

\vspace{6pt}
\tcp{Student}\vspace{2.5pt}
\If{step \% ratio == 0}{
\vspace{3pt}
\assignline[eq:dual_dmd_grad]{$\mathcal{L}^{\text{trg}+\text{src}}_\text{DMD}$}
                            {$\text{DualDMD}~\big(x_t,\epsilon,a\big)$}
                            \vspace{3pt}
\assignline[eq:gan_g_loss]{$\mathcal{L}_{\text{GAN}}^{G}$}{$\text{MhGAN}~\big(x_t\big)$}\vspace{2pt}
\assignline[eq:g_total_loss]{$\mathcal{L}_{G}$}
                      {$\mathcal{L}^{\text{trg}+\text{src}}_\text{DMD}+\lambda_{\text{GAN}}^{G}\,\mathcal{L}_{\text{GAN}}^{G}$}
                      \vspace{3pt}
\assignline{$G$}{$\text{update}~\big(G,\mathcal{L}_{G}\big)$}
}
\vspace{6pt}
\tcp{Fake Teacher \& Discriminator}\vspace{2.5pt}
\assignline[eq:fake_loss]{$\mathcal{L}_{\text{fk}}$}{$\text{MSE}~~\!\big(\epsilon^\text{fk}(\text{stop\_grad}(x_t)),\,\epsilon\big)$}
\vspace{3pt}
\assignline[eq:gan_d_loss]{$\mathcal{L}_{\text{GAN}}^{D}$}
                        {$\text{MhGAN}~\big(\text{stop\_grad}(x_t),y_t\big)$}
                        \vspace{3pt}
\assignline[eq:fk_total_loss]{$\mathcal{L}_{\text{fk}+D}$}
{$\mathcal{L}_{\text{fk}} + \lambda_{\text{GAN}}^{D}\,\mathcal{L}_{\text{GAN}}^{D}$}
\vspace{3pt}
\assignline{$\epsilon^\text{fk}$}{$\text{update}~\big(\epsilon^\text{fk},\mathcal{L}_{\text{fk}+D}\big)$}

\vspace{6pt}
\tcp{Target Teacher}\vspace{2.5pt}
\If{step \% ratio == 0 ~\&~ \text{train\_target}}{
\assignline[eq:target_loss]{$\mathcal{L}_{\epsilon^\text{trg}}$}{$\text{MSE}~\!\big(\epsilon^\text{trg}(y_t),\,\epsilon\big)$}
\assignline{$\epsilon^\text{trg}$}{$\text{update}~\big(\epsilon^\text{trg},\mathcal{L}_{\text{trg}}\big)$}
}

\end{algorithm}

%% file: src/tables/4_FSIG_kshot.tex
\begin{table}[t]
\centering\footnotesize
\caption{Ablation on target set sizes and $\text{NFE}$, evaluated by \fid. B: Babies, M: MetFaces. \textbf{Bold} indicates best result. Selected variant for main results is in \colorbox{black!5}{gray}.}
\label{tab:4_kshot_fid}
\setlength{\tabcolsep}{3.5pt}
\begin{tabularx}{\columnwidth}{l c *{6}{c}}
\toprule
& & \multicolumn{2}{c}{\textbf{1-shot}} & \multicolumn{2}{c}{\textbf{5-shot}} & \multicolumn{2}{c}{\textbf{10-shot}} \\
\cmidrule(lr){3-4}\cmidrule(lr){5-6}\cmidrule(lr){7-8}
\textbf{Methods} & \textbf{NFE$\downarrow$} & B & M & B & M & B & M \\
\midrule
CRDI & 25 & 105.51 & 145.10 & 51.71 & 126.34 & 48.52 & 121.36 \\
\addlinespace
\texttt{Uni-DAD} & 4 & \textbf{72.38} & 95.44 & \textbf{45.86} & \textbf{81.85} & \textbf{41.39} & 59.49 \\
\texttt{Uni-DAD} & 3 & 90.33 & \textbf{90.29} & 52.73 & 83.69 & \cellcolor{black!5} 45.09 & \cellcolor{black!5} \textbf{58.13} \\
\texttt{Uni-DAD} & 2 & 95.69 & 114.78 & 68.75 & 98.36 & 62.45 & 79.08 \\
\texttt{Uni-DAD} & 1 & 109.55 & 132.79 & 93.52 & 103.84 & 98.52 & 89.03 \\

\bottomrule
\end{tabularx}
\end{table}

%% file: src/tables/4_FSIG_Ablation_Components.tex
\begin{table}[t]
\centering\footnotesize
\caption{Component analysis evaluated by 
\fid.
\label{tab:4_component_ablation}
Mh: Multi-head, Sh: Single-head, B: Babies, M: MetFaces.
 \textbf{Bold} indicates best result. Selected variants for main results are in \colorbox{black!5}{gray}. }
\label{tab:4_components}
\setlength{\tabcolsep}{3pt}
\renewcommand{\arraystretch}{-1} 

\begin{tabularx}{\columnwidth}{@{}%
  Y
  C{0.115\columnwidth} C{0.115\columnwidth} C{0.115\columnwidth} C{0.115\columnwidth}%
  C{0.10\columnwidth} C{0.10\columnwidth}
@{}}
\toprule
\multirow{2}{*}{\textbf{Group}} &
\multicolumn{4}{>{\columncolor{white}}c}{\textbf{Components}} &
\multicolumn{2}{c}{\textbf{FID} $\downarrow$} \\
\cmidrule(lr){2-5}\cmidrule(l){6-7}
& DMD$^{\text{src}}$
& DMD$^{\text{trg}}$
& GAN$^{\text{Mh}}$
& GAN$^{\text{Sh}}$
& B & M \\
\midrule
\textit{GAN\_only} &      &        &        & \cmark & 56.90  & 80.14 \\
\arrayrulecolor{black!10}\cmidrule(l{0pt}r{0pt}){2-5}\arrayrulecolor{black}
  &        &        & \cmark &        & 130.34 & 110.00 \\
\midrule

\textit{DMD\_only} &      &   \cmark     &        & & 110.39 & 68.05 \\
\arrayrulecolor{black!10}\cmidrule(l{0pt}r{0pt}){2-5}\arrayrulecolor{black}
  &      \cmark    &       & &        & 166.99 & 105.67 \\
\arrayrulecolor{black!10}\cmidrule(l{0pt}r{0pt}){2-5}\arrayrulecolor{black}
  & \cmark & \cmark &        &        & 84.19  & 69.80 \\
  \midrule

\textit{DMD+GAN}      &  & \cmark &        & \cmark & 54.18  & 68.54 \\
\arrayrulecolor{black!10}\cmidrule(l{0pt}r{0pt}){2-5}\arrayrulecolor{black}
    & \cmark &  &        & \cmark &  57.58 &  83.67\\
\arrayrulecolor{black!10}\cmidrule(l{0pt}r{0pt}){2-5}\arrayrulecolor{black}
  & \cmark & \cmark &        & \cmark & 48.89  & 80.24 \\
\arrayrulecolor{black!10}\cmidrule(l{0pt}r{0pt}){2-5}\arrayrulecolor{black}
&        & \cmark & \cmark &        & 54.68  & 68.74 \\
\arrayrulecolor{black!10}\cmidrule(l{0pt}r{0pt}){2-5}\arrayrulecolor{black}
\textit{}& \cmark &        & \cmark &        & \cellcolor{black!5}47.38  & \cellcolor{black!5}64.13 \\
\arrayrulecolor{black!10}\cmidrule(l{0pt}r{0pt}){2-5}\arrayrulecolor{black}

& \cmark & \cmark & \cmark &        & \cellcolor{black!5}\textbf{45.09} & \cellcolor{black!5}\textbf{58.13} \\
\bottomrule
\end{tabularx}
\end{table}

%% file: src/tables/4_FSIG_Ablation_Init.tex
\begin{table}[t]
\centering\footnotesize
\caption{Available checkpoints at the start of training and \fid. $G$ is distilled via DMD2~\cite{yin2024improved} and $\epsilon^{\text{trg}}$ is adapted via fine-tuning. \textbf{Bold} indicates best result. Selected variant for main results is in \colorbox{black!5}{gray}.} 
\label{tab:4_init_ablation}
\setlength{\tabcolsep}{4pt} 
\renewcommand{\arraystretch}{-1} 
\begin{tabularx}{0.92\columnwidth}{@{} Y Y Y Y @{}}
\toprule
\multicolumn{2}{c}{\textbf{Available Checkpoints}} &
\multicolumn{2}{c}{\textbf{FID} $\downarrow$} \\
\cmidrule(lr){1-2}\cmidrule(lr){3-4}
\textit{Pre-distilled $G$} &
\mbox{\textit{Pre-adapted $\epsilon^{\text{trg}}$}} & Babies & MetFaces \\
\midrule
      &       & \cellcolor{black!5} 45.09 & 
      \cellcolor{black!5}58.13 \\
\arrayrulecolor{black!10}\cmidrule(l{0pt}r{0pt}){1-2}\arrayrulecolor{black}
\cmark&       & 42.68 & 77.80 \\
\arrayrulecolor{black!10}\cmidrule(l{0pt}r{0pt}){1-2}\arrayrulecolor{black}
      & \cmark&  46.73 & 62.67 \\
\arrayrulecolor{black!10}\cmidrule(l{0pt}r{0pt}){1-2}\arrayrulecolor{black}
\cmark & \cmark & \textbf{42.04} & \textbf{54.01} \\
\bottomrule
\end{tabularx}
\end{table}

%% file: src/tables/4_FSIG_Ablation_GANloss.tex
\begin{table}[t]
\centering\footnotesize
\caption{Ablation of GAN losses evaluated on \fid. \textbf{Bold} indicates best result. Selected variant for main results is in \colorbox{Black!5}{gray}.}
\label{tab:4_gan_loss_ablation}
\setlength{\tabcolsep}{4pt}
\begin{tabularx}{\columnwidth}{@{} l *{4}{>{\centering\arraybackslash}X} @{}}
\toprule
 & \multicolumn{2}{c}{\textbf{Multi-head}} & \multicolumn{2}{c}{\textbf{Single-head}} \\
\cmidrule(lr){2-3}\cmidrule(lr){4-5}
\textbf{GAN Loss} & Babies & MetFaces & Babies & MetFaces \\
\midrule
BCE   & \cellcolor{black!5}\textbf{45.09} & \cellcolor{black!5}\textbf{58.13} &  48.89 & 80.24 \\
Hinge  & 45.82 & 64.83 & 50.92  & 72.11 \\
LSGAN  & 46.18 & 91.55 & \cellcolor{black!5}\textbf{47.98} & \cellcolor{black!5}\textbf{64.36} \\
WGAN   & 64.54 & 95.35 & 58.50 & 84.91 \\
\bottomrule
\end{tabularx}
\end{table}

%% file: src/tables/4_SDIG_qualitative_diversity.tex
\begin{figure*}[!t]
\centering\footnotesize

\setlength{\tabcolsep}{2pt}

\newcommand{\doggtarget}{\includegraphics[width=\colw,height=\colw,keepaspectratio]{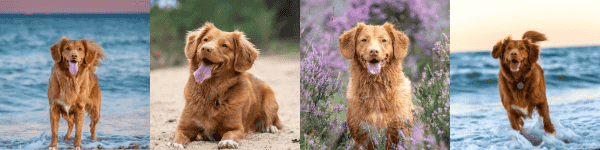}}

\newcommand{\dogunidadmountain}{\includegraphics[width=\colw,height=\colw,keepaspectratio]{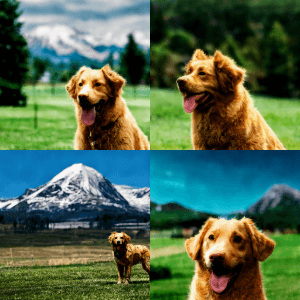}}
\newcommand{\dogunidadinter}{\includegraphics[width=\colw,height=\colw,keepaspectratio]{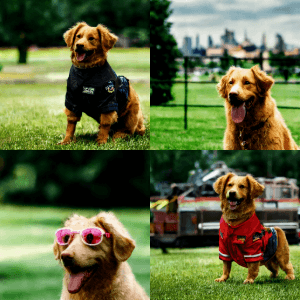}}

\newcommand{\dogdbmountain}{\includegraphics[width=\colw,height=\colw,keepaspectratio]{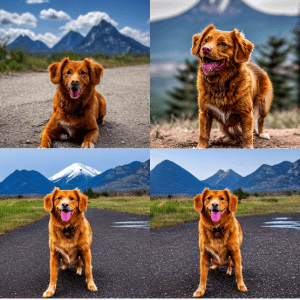}}
\newcommand{\dogdbinter}{\includegraphics[width=\colw,height=\colw,keepaspectratio]{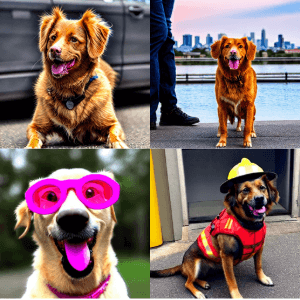}}

\newcommand{\dogpsomountain}{\includegraphics[width=\colw,height=\colw,keepaspectratio]{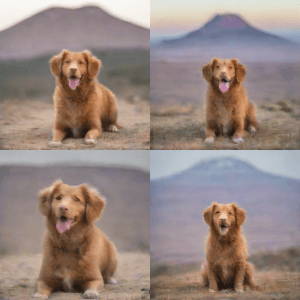}}
\newcommand{\dogpsointer}{\includegraphics[width=\colw,height=\colw,keepaspectratio]{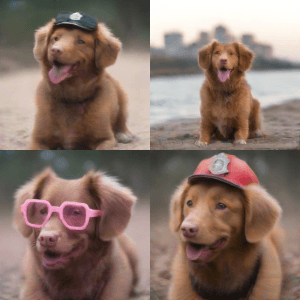}}

\newcommand{\dogdbdmdmountain}{\includegraphics[width=\colw,height=\colw,keepaspectratio]{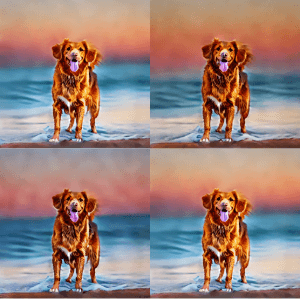}}
\newcommand{\dogdbdmdinter}{\includegraphics[width=\colw,height=\colw,keepaspectratio]{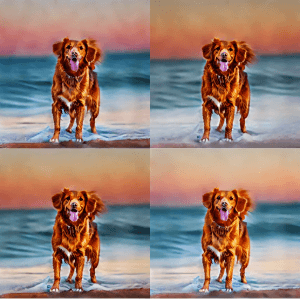}}

\newcommand{\dogdmddbmountain}{\includegraphics[width=\colw,height=\colw,keepaspectratio]{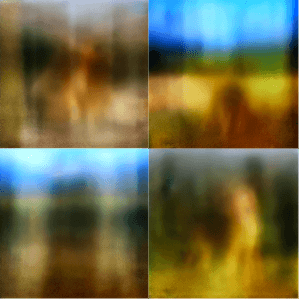}}
\newcommand{\dogdmddbinter}{\includegraphics[width=\colw,height=\colw,keepaspectratio]{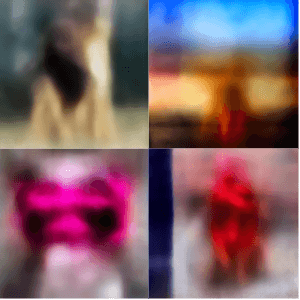}}

\begingroup
\newcommand{\headsize}{\footnotesize}

\setlength{\arrayrulewidth}{0.8pt}%
\arrayrulecolor{black}%

\resizebox{\textwidth}{!}{
\begin{tabular}{
  *{5}{>{\centering\arraybackslash}m{\colw}}
}

\shortstack[l]{\headsize\bfseries Target Set\\\headsize ``a \texttt{prt} dog''} &
\doggtarget & & & \\[2pt]
\cmidrule(lr){1-5}

\shortstack[c]{\headsize\bfseries \outlinebox{~\texttt{Uni-DAD}~}\\\headsize SDv1.5, NFE = 1} &
\shortstack[c]{\headsize\bfseries FT\\\headsize SDv1.5, NFE = 2$\times$50} &
\shortstack[c]{\headsize\bfseries \textcolor{red}{DMD2-FT}\\\headsize SDv1.5, NFE = 1} &
\shortstack[c]{\headsize\bfseries \textcolor{orange}{FT-DMD2}\\\headsize SDv1.5, NFE = 1} &
\shortstack[c]{\headsize\bfseries Turbo-PSO\\\headsize SDXL, NFE = 4} \\[2pt]
\cmidrule(lr){1-5}

\multicolumn{5}{l}{\textbf{Same Prompt:} ``a \texttt{prt} dog with a mountain in the background''} \\[3pt]

\dogunidadmountain & \dogdbmountain & \dogdmddbmountain & \dogdbdmdmountain & \dogpsomountain \\
\\[-7pt]
\cmidrule(lr){1-5}

\multicolumn{5}{l}{%
\parbox{0.95\textwidth}{%
\textbf{Different Prompts:} ``a \texttt{prt} dog... \textbf{a:} in a police outfit'' \textbf{b:} with a city skyline in the background'' \textbf{c:} with pink sunglasses'' \textbf{d:} in a firefighter outfit.''}} \\[3pt]

\dogunidadinter & \dogdbinter & \dogdmddbinter & \dogdbdmdinter & \dogpsointer \\

\end{tabular}
} 
\endgroup

\caption{Qualitative \textbf{diversity} comparison for SDP, adapting SDv1.5~\cite{rombach2022high} to the DreamBooth~\cite{ruiz2023dreambooth} \emph{dog7} subject. The first row shows generations under the same prompt (Intra-LPIPS logic). The second row shows generations across different prompts (Inter-LPIPS logic). \textbf{a:} top-left, \textbf{b:} top-right, \textbf{c:} bottom-left, \textbf{d:} bottom-right.}
\label{fig:diversity_qualitative}

\end{figure*}

%% file: src/tables/4_SDIG_qualitative.tex
\begin{figure*}[t]
\centering\footnotesize

\setlength{\tabcolsep}{2pt}

\newcommand{\cattarget}{\includegraphics[width=\colw,height=\colw,keepaspectratio]{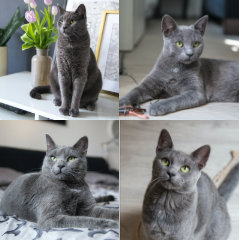}}

\newcommand{\dogtarget}{\includegraphics[width=\colw,height=\colw,keepaspectratio]{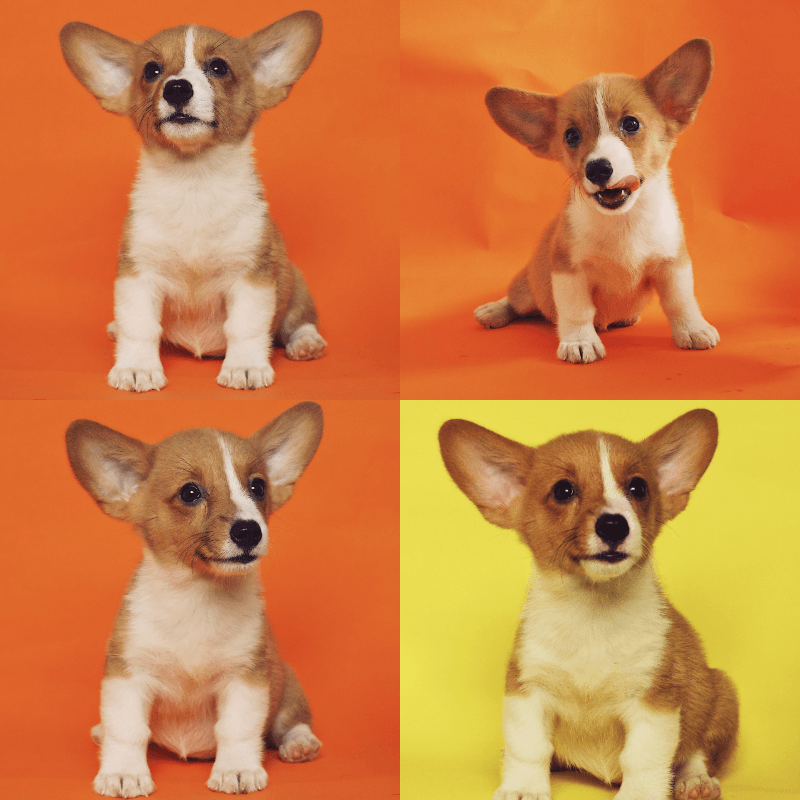}}

\newcommand{\vasetarget}{\includegraphics[width=\colw,height=\colw,keepaspectratio]{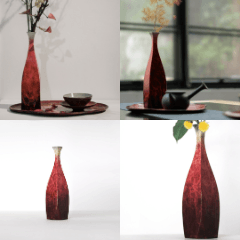}}

\newcommand{\dogunidadmountain}{\includegraphics[width=\colw,height=\colw,keepaspectratio]{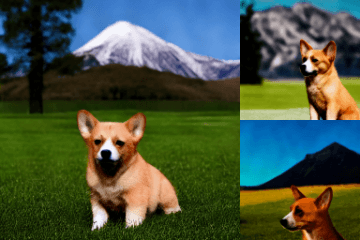}}
\newcommand{\dogunidadblue}{\includegraphics[width=\colw,height=\colw,keepaspectratio]{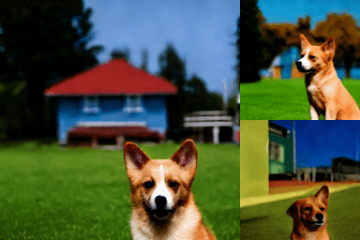}}
\newcommand{\dogunidadsnow}{\includegraphics[width=\colw,height=\colw,keepaspectratio]{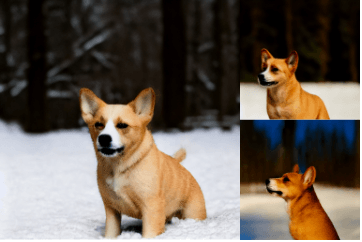}}

\newcommand{\catunidadpinkg}{\includegraphics[width=\colw,height=\colw,keepaspectratio]{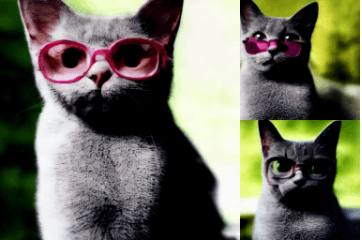}}
\newcommand{\catunidadjungle}{\includegraphics[width=\colw,height=\colw,keepaspectratio]{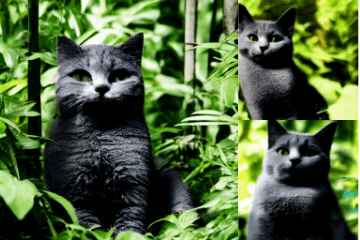}}
\newcommand{\catunidadfiref}{\includegraphics[width=\colw,height=\colw,keepaspectratio]{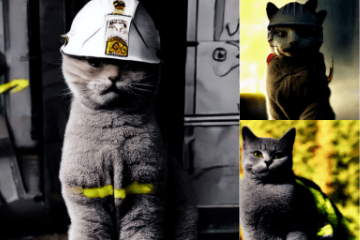}}

\newcommand{\vaseunidadbeach}{\includegraphics[width=\colw,height=\colw,keepaspectratio]{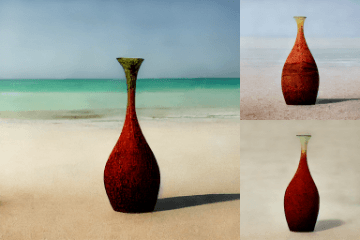}}
\newcommand{\vaseunidadsunf}{\includegraphics[width=\colw,height=\colw,keepaspectratio]{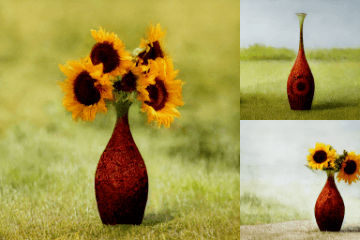}}
\newcommand{\vaseunidadcob}{\includegraphics[width=\colw,height=\colw,keepaspectratio]{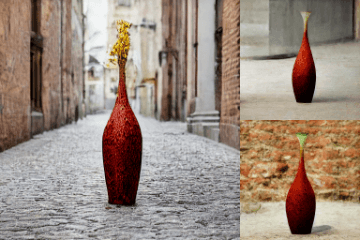}}

\newcommand{\dogdbmountain}{\includegraphics[width=\colw,height=\colw,keepaspectratio]{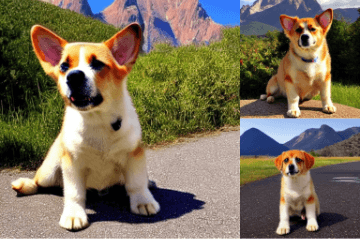}}
\newcommand{\dogdbblue}{\includegraphics[width=\colw,height=\colw,keepaspectratio]{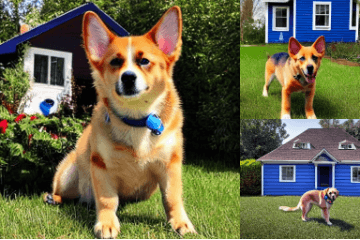}}
\newcommand{\dogdbsnow}{\includegraphics[width=\colw,height=\colw,keepaspectratio]{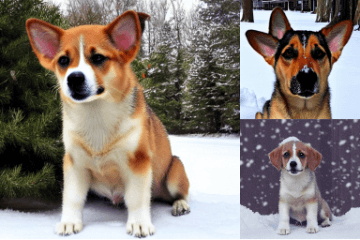}}

\newcommand{\vasedbbeach}{\includegraphics[width=\colw,height=\colw,keepaspectratio]{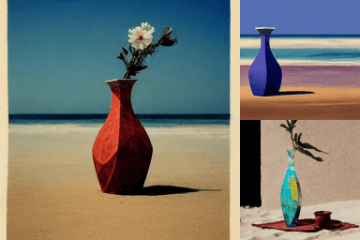}}
\newcommand{\vasedbcobb}{\includegraphics[width=\colw,height=\colw,keepaspectratio]{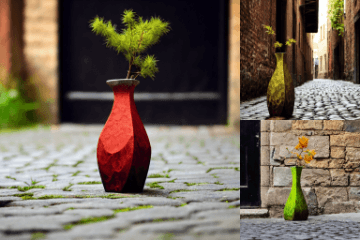}}
\newcommand{\vasedbsunf}{\includegraphics[width=\colw,height=\colw,keepaspectratio]{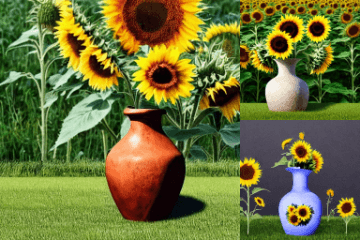}}

\newcommand{\catdbjungle}{\includegraphics[width=\colw,height=\colw,keepaspectratio]{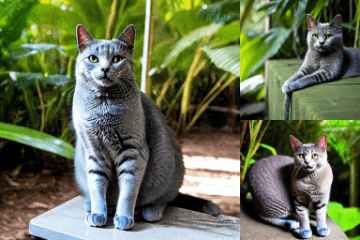}}
\newcommand{\catdbpink}{\includegraphics[width=\colw,height=\colw,keepaspectratio]{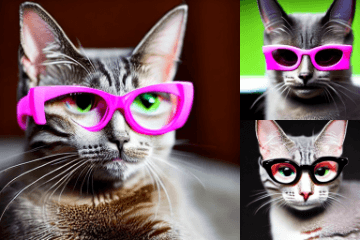}}
\newcommand{\catdbfiref}{\includegraphics[width=\colw,height=\colw,keepaspectratio]{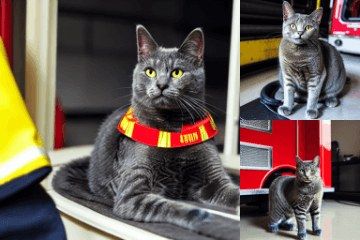}}

\newcommand{\dogddbmountain}{\includegraphics[width=\colw,height=\colw,keepaspectratio]{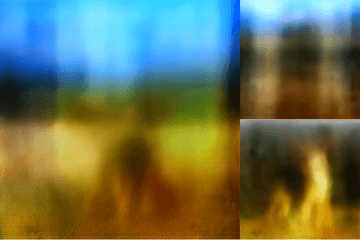}}
\newcommand{\dogddbblue}{\includegraphics[width=\colw,height=\colw,keepaspectratio]{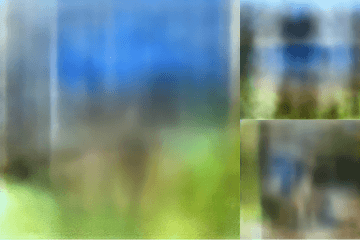}}
\newcommand{\dogddbsnow}{\includegraphics[width=\colw,height=\colw,keepaspectratio]{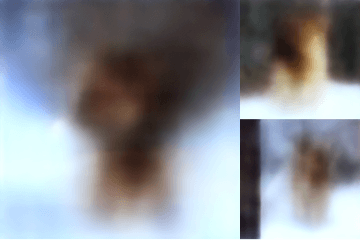}}

\newcommand{\catddbpink}{\includegraphics[width=\colw,height=\colw,keepaspectratio]{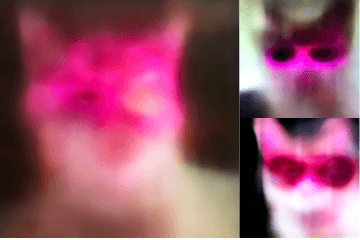}}
\newcommand{\catddbjungle}{\includegraphics[width=\colw,height=\colw,keepaspectratio]{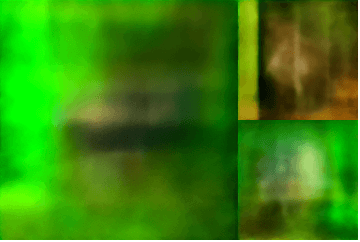}}
\newcommand{\catddbfiref}{\includegraphics[width=\colw,height=\colw,keepaspectratio]{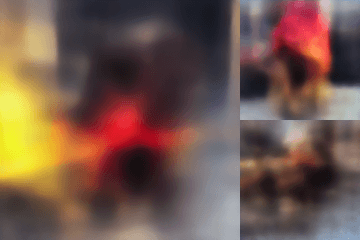}}

\newcommand{\vaseddbbeach}{\includegraphics[width=\colw,height=\colw,keepaspectratio]{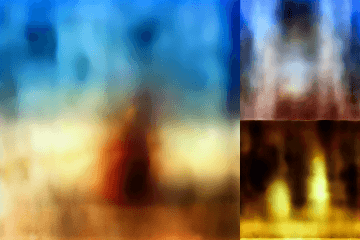}}
\newcommand{\vaseddbcobb}{\includegraphics[width=\colw,height=\colw,keepaspectratio]{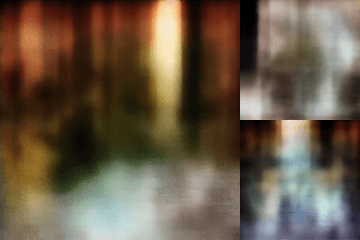}}
\newcommand{\vaseddbsunf}{\includegraphics[width=\colw,height=\colw,keepaspectratio]{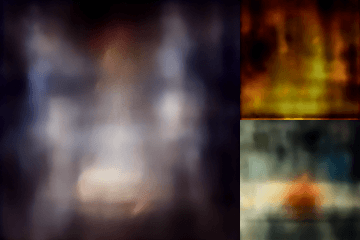}}

\newcommand{\vasedbdmdbeach}{\includegraphics[width=\colw,height=\colw,keepaspectratio]{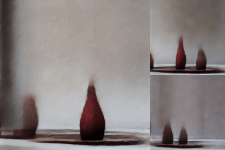}}
\newcommand{\vasedbdmdsun}{\includegraphics[width=\colw,height=\colw,keepaspectratio]{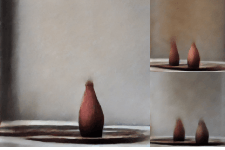}}
\newcommand{\vasedbdmdstreet}{\includegraphics[width=\colw,height=\colw,keepaspectratio]{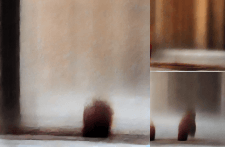}}

\newcommand{\catdpinkdbdmd}{\includegraphics[width=\colw,height=\colw,keepaspectratio]{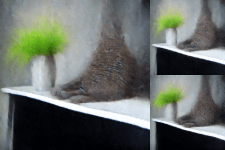}}
\newcommand{\catjungledbdmd}{\includegraphics[width=\colw,height=\colw,keepaspectratio]{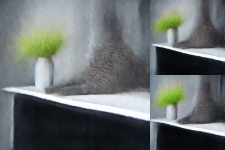}}
\newcommand{\catfirfdbdmd}{\includegraphics[width=\colw,height=\colw,keepaspectratio]{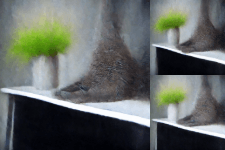}}

\newcommand{\dogmountaindbdmd}{\includegraphics[width=\colw,height=\colw,keepaspectratio]{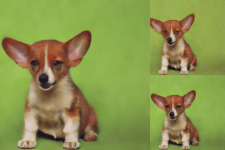}}
\newcommand{\dogsnowdbdmd}{\includegraphics[width=\colw,height=\colw,keepaspectratio]{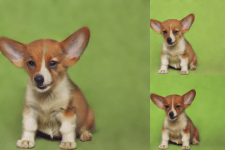}}
\newcommand{\dogbluedbdmd}{\includegraphics[width=\colw,height=\colw,keepaspectratio]{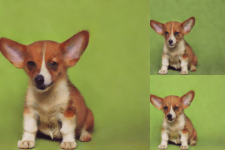}}

\newcommand{\dogpsomountain}{\includegraphics[width=\colw,height=\colw,keepaspectratio]{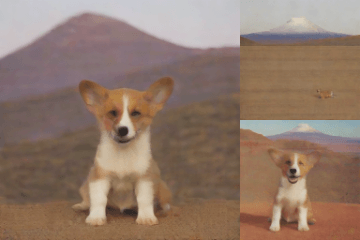}}
\newcommand{\dogpsoblue}{\includegraphics[width=\colw,height=\colw,keepaspectratio]{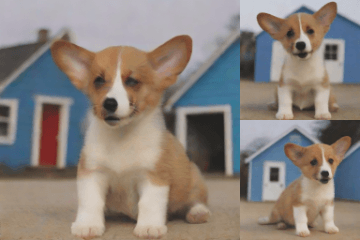}}
\newcommand{\dogpsosnow}{\includegraphics[width=\colw,height=\colw,keepaspectratio]{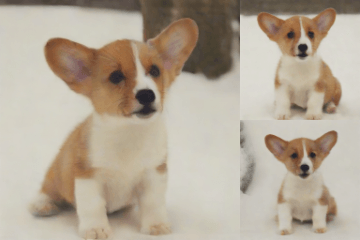}}

\newcommand{\catpsopink}{\includegraphics[width=\colw,height=\colw,keepaspectratio]{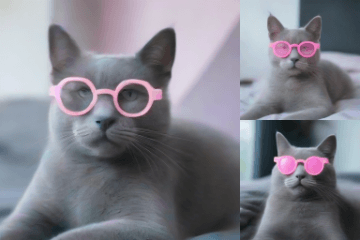}}
\newcommand{\catpsojungle}{\includegraphics[width=\colw,height=\colw,keepaspectratio]{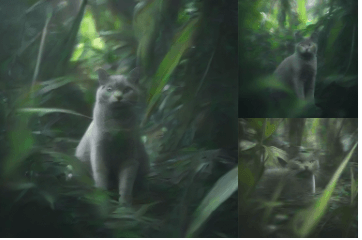}}
\newcommand{\catpsofiref}{\includegraphics[width=\colw,height=\colw,keepaspectratio]{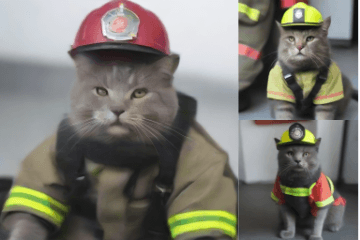}}

\newcommand{\vasepsobeach}{\includegraphics[width=\colw,height=\colw,keepaspectratio]{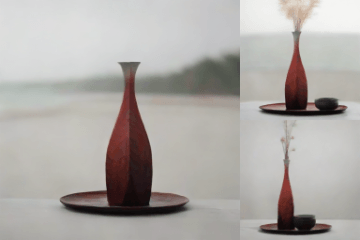}}
\newcommand{\vasepsocobb}{\includegraphics[width=\colw,height=\colw,keepaspectratio]{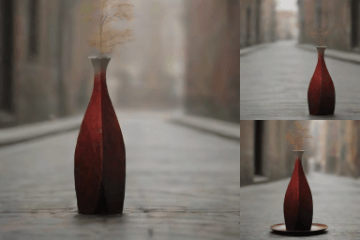}}
\newcommand{\vasepsosunf}{\includegraphics[width=\colw,height=\colw,keepaspectratio]{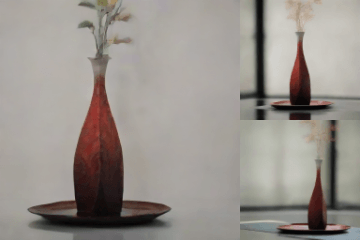}}

\newcommand{\backpackpso}{\includegraphics[width=\colw,height=\colw,keepaspectratio]{src/figures/picture_SDIG/backpack_pso.png}}

\newcommand{\wolfypso}{\includegraphics[width=\colw,height=\colw,keepaspectratio]{src/figures/picture_SDIG/wolfy_pso.png}}

\begingroup
\newcommand{\headsize}{\footnotesize}
\newcommand{\hdrplain}[1]{\makebox[\colw][c]{\headsize\bfseries #1}}

\newcommand{\threeprompt}[1]{\multicolumn{3}{l}{\scriptsize #1}}

\begingroup
\setlength{\arrayrulewidth}{0.8pt}%
\arrayrulecolor{black}%

\resizebox{0.90\textwidth}{!}{
\begin{tabular}{
  >{\centering\arraybackslash}m{\colw} 
  >{\centering\arraybackslash}m{\colw} 
  *{5}{>{\centering\arraybackslash}m{\colw}}
}
\hdrplain{Target Set} &
\shortstack[c]{\headsize\bfseries \outlinebox{~\texttt{Uni-DAD}~}\\\headsize SDv1.5, NFE = 1} &
\shortstack[c]{\headsize\bfseries FT\\\headsize SDv1.5, NFE = 2$\times$50} &
\shortstack[c]{\headsize\bfseries \textcolor{red}{DMD2-FT}\\\headsize SDv1.5, NFE = 1} &
\shortstack[c]{\headsize\bfseries \textcolor{orange}{FT-DMD2}\\\headsize SDv1.5, NFE = 1} &
\shortstack[c]{\headsize\bfseries Turbo-PSO\\\headsize SDXL, NFE = 4} \\[2pt]
\midrule

 & \dogunidadmountain & \dogdbmountain & \dogddbmountain & \dogmountaindbdmd & \dogpsomountain  \\
{\scriptsize    } & \threeprompt{``a \texttt{prt} dog with a mountain in the background''} & \\[-7pt]
\dogtarget& \dogunidadsnow& \dogdbsnow & \dogddbsnow & \dogsnowdbdmd & \dogpsosnow  \\[-10pt]
{\scriptsize } & \threeprompt{``a \texttt{prt} dog in the snow''}& \\[+3pt]
{\scriptsize ``a \texttt{prt} dog''} & \dogunidadblue & \dogdbblue & \dogddbblue & \dogbluedbdmd & \dogpsoblue  \\[+4pt]
{\scriptsize    } &\threeprompt{``a \texttt{prt} dog with a blue house in the background''}& \\[+3pt]
\midrule

\ & \catunidadpinkg & \catdbpink & \catddbpink & \catdpinkdbdmd & \catpsopink  \\
{\scriptsize    } &\threeprompt{``a \texttt{prt} cat wearing pink glasses''} \\[-7pt]
\cattarget & \catunidadjungle & \catdbjungle & \catddbjungle & \catjungledbdmd & \catpsojungle  \\[-10pt]
{\scriptsize } &\threeprompt{``a \texttt{prt} cat in the jungle''} &\\[+3pt]
{\scriptsize ``a \texttt{prt} cat''} & \catunidadfiref & \catdbfiref & \catddbfiref & \catfirfdbdmd & \catpsofiref  \\[+4pt]
{\scriptsize    } &\threeprompt{``a \texttt{prt} cat in a firefighter outfit''} &\\[+3pt]

\midrule

\ & \vaseunidadbeach & \vasedbbeach & \vaseddbbeach & \vasedbdmdbeach & \vasepsobeach   \\
{\scriptsize    } &\threeprompt{``a \texttt{prt} vase on the beach''} \\[-7pt]
\vasetarget & \vaseunidadsunf & \vasedbsunf & \vaseddbsunf & \vasedbdmdsun & \vasepsosunf  \\[-10pt]
{\scriptsize } &\threeprompt{``a \texttt{prt} vase on top of green grass with sunflowers around it''} &\\[+3pt]
{\scriptsize ``a \texttt{prt} vase''} & \vaseunidadcob & \vasedbcobb & \vaseddbcobb & \vasedbdmdstreet & \vasepsocobb  \\[+4pt]
{\scriptsize    } &\threeprompt{``a \texttt{prt} vase on a cobblestone street''} &\\[+3pt]

\end{tabular}
} 
\arrayrulecolor{black}%
\endgroup

\caption{Continued from Fig.~\ref{fig:4_SDP_qualitative}: Qualitative comparison for SDP, adapting SDV1.5~\cite{rombach2022high} to the DreamBooth~\cite{ruiz2023dreambooth} \emph{cat2, dog6, vase} subjects, evaluated on accessorization and re-contextualization prompts. Zoom in for details.}
\label{fig:A_SDP_qualitative}
\endgroup
\end{figure*}